\documentclass[default]{sn-jnl}%

\usepackage{graphicx}%
\usepackage{svg}
\usepackage{multirow}%
\usepackage{amsmath,amssymb,amsfonts}%
\usepackage{array}
\usepackage[title]{appendix}%
\usepackage{xcolor}%
\usepackage{manyfoot}%
\usepackage{booktabs}%
\usepackage{listings}%
\usepackage[title]{appendix}
\usepackage{setspace}

\usepackage{mathtools}
\usepackage{makecell}
\usepackage[11pt]{moresize}
\usepackage{tabularx}
\usepackage{rotating}
\usepackage{subcaption}
\usepackage{adjustbox}

\usepackage{pifont}%

\usepackage[titlenumbered, ruled, linesnumbered, noend]{algorithm2e}
\SetAlFnt{\scriptsize} %
\SetInd{.2cm}{.2cm}
\SetArgSty{textnormal}
\SetKwComment{tcp}{~~~$\blacktriangleright$ }{}
\SetKwComment{tcpline}{$\blacktriangledown$ }{}

\SetCommentSty{mycommfont}
\SetKw{And}{and}
\SetKw{Break}{break}
\SetKw{Or}{or}
\SetKw{In}{in}
\SetKw{Continue}{continue}
\let\oldnl\nl%
\newcommand{\nonl}{\renewcommand{\nl}{\let\nl\oldnl}}
\AtBeginEnvironment{algorithm}{\SetProgSty{textrm}\SetArgSty{textrm}\SetFuncSty{textsc}\SetFuncArgSty{textrm}}

\usepackage{xspace}

\usepackage{orcidlink}
\usepackage{enumitem}
\usepackage{amsmath}
\usepackage{stmaryrd}
\usepackage{dsfont}
\DeclareMathAlphabet{\mathpzc}{OT1}{pzc}{m}{it}
\usepackage{calligra}
\DeclareMathAlphabet{\mathcalligra}{T1}{calligra}{m}{n}
\usepackage{verbatim}
\usepackage{upgreek}
\usepackage{bbm}

\usepackage{cellspace}
\newcolumntype{L}[1]{>{\raggedright\arraybackslash} S{m{#1}} }

\makeatletter
\newcommand\incircbin
{%
  \mathpalette\@incircbin
}
\newcommand\@incircbin[2]
{%
  \mathbin%
  {%
    \ooalign{\hidewidth$#1#2$\hidewidth\crcr$#1\bigcirc$}%
  }%
}

\makeatother

\makeatletter
\DeclareRobustCommand\bigop[2][1]{%
  \mathop{\vphantom{\sum}\mathpalette\bigop@{{#1}{#2}}}\slimits@
}
\newcommand{\bigop@}[2]{\bigop@@#1#2}
\newcommand{\bigop@@}[3]{%
  \vcenter{%
    \sbox\z@{$#1\sum$}%
    \hbox{\resizebox{\ifx#1\displaystyle#2\fi\dimexpr\ht\z@+\dp\z@}{!}{$\m@th#3$}}%
  }%
}
\let\oldin\in
\renewcommand{\in}{{\,\oldin\,}}

\newcommand{\definition}[1]{\textbf{#1}}

\newcommand{\locomotifdok}{\mbox{LoCoMotif-DoK}\xspace}

\newcommand{\mset}{\mathcal{M}} %
\newcommand{\cmset}{\mathcal{C}} %
    \newcommand{\hard}{H}
    \newcommand{\desir}{D}
    \newcommand{\hmot}{\hard^\text{mot}}
    \newcommand{\hmotssame}{\hard^\text{mots-same}}
    \newcommand{\hmotsdiff}{\hard^\text{mots-diff}}
    \newcommand{\hmset}{\hard^\text{mset}}
    \newcommand{\hmsets}{\hard^\text{msets}}
    \newcommand{\dmset}{\desir} %
    \newcommand{\hmotrepr}{\hard^\text{mot-repr}}
    \newcommand{\Hard}{H}
    \newcommand{\Desir}{D}
    \newcommand{\Hmot}{\Hard^\text{mot}}
    \newcommand{\Hmotssame}{\Hard^\text{mots-same}}
    
    \newcommand{\Hmset}{\Hard^\text{mset}}
    \newcommand{\Hmsets}{\Hard^\text{msets}}
    \newcommand{\Dmset}{\Desir} %
    \newcommand{\Hmotrepr}{\Hard^\text{mot-repr}}
    \newcommand{\HARD}{\mathbf{H}}
    \newcommand{\DESIR}{\mathbf{D}}

    \newcommand{\HMSET}{\HARD^\text{mset}}
    \newcommand{\HMSETS}{\HARD^\text{msets}}
    \newcommand{\DMSET}{\DESIR} %

\DeclareMathOperator*{\avg}{avg}

\DeclareMathOperator*{\argmax}{arg\, max}

\usepackage{natbib}
\setcitestyle{authoryear}
\setcitestyle{round}
\setcitestyle{semicolon}
\setcitestyle{aysep={}}

\usepackage[textsize=scriptsize]{todonotes}
\usepackage{marginnote}

\usepackage{cancel}
\usepackage[normalem]{ulem}
\usepackage{soul}
\usepackage{xcolor}

\begin{document}

\title{Steering the LoCoMotif: Using Domain Knowledge in Time Series Motif Discovery}

\author*[1,2]{\mbox{\fnm{Aras} \sur{Yurtman} \orcidlink{0000-0001-6213-5427}}}\email{aras.yurtman@kuleuven.be}
\author[1,2]{\mbox{\fnm{Daan} \sur{Van Wesenbeeck} \orcidlink{0000-0002-4941-5480}}}
\author[1,2]{\mbox{\fnm{Wannes} \sur{Meert} \orcidlink{0000-0001-9560-3872}}}
\author[1,2]{\mbox{\fnm{Hendrik} \sur{Blockeel} \orcidlink{0000-0003-0378-3699}}}

\affil[1]{\orgdiv{Dept. of Computer Science}, \orgname{KU Leuven}, \orgaddress{\city{Leuven}, \postcode{B-3000}, \country{Belgium}}}
\affil[2]{\orgdiv{Leuven.AI - KU Leuven Institute for AI}, \city{Leuven}, \postcode{B-3000}, \country{Belgium}}

\abstract{
Time Series Motif Discovery (TSMD) identifies repeating patterns in time series data, but its unsupervised nature might result in motifs that are not interesting to the user. 
To address this, we propose a framework that allows the user to impose constraints on the motifs to be discovered, 
where constraints can easily be defined 
according to the properties of the desired motifs in the application domain.
We also propose an efficient implementation of the framework, the \locomotifdok algorithm. 
We demonstrate that \locomotifdok can effectively leverage domain knowledge in real and synthetic data, outperforming other TSMD techniques which only support a limited form of domain knowledge.
}

\maketitle

\section{Introduction}\label{sec:introduction}

Time series data often contain repeating patterns, some of which might be interesting because they reveal important aspects of the underlying process. 
Such patterns can be manually extracted by experts, but this is a time-consuming and costly approach. 
An alternative is to use \definition{Time Series Motif Discovery (TSMD)}, which automatically identifies (approximately) repeating patterns in a time series in an unsupervised manner.
In the context of TSMD, a time segment that contains a single occurrence of a pattern is called a \definition{motif}, and the set of all occurrences of the same pattern is called a \definition{motif set}.

\begin{figure}
    \centering
    % \includesvg[width=\textwidth]{figures/physical_therapy_example_paper_2_4_fig.svg}
    \includegraphics[width=\textwidth]{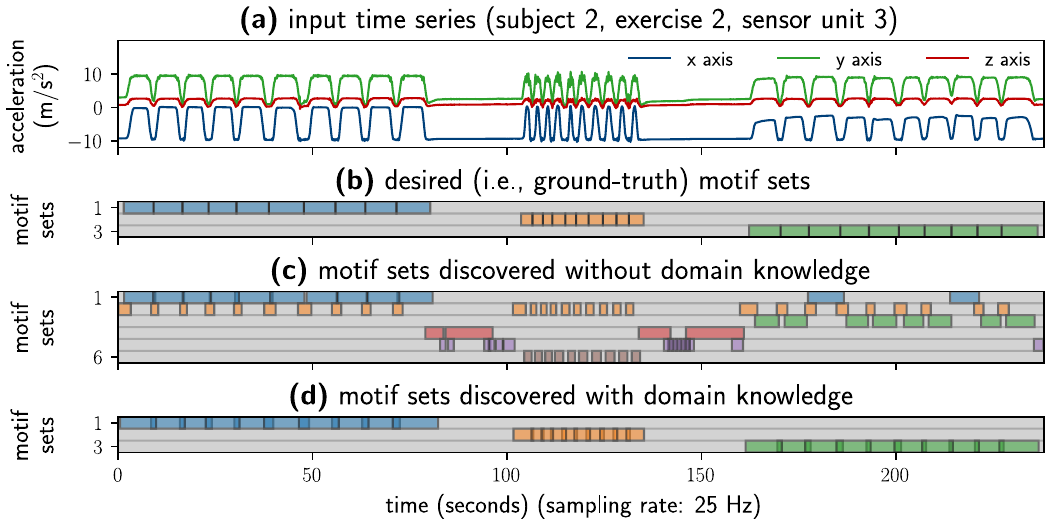}
    \caption{%
    (a)~From the time series that captures the acceleration of the lower arm, (b)~executions of a physical therapy exercise are (d)~successfully discovered by using domain knowledge, (c)~but not without using domain knowledge. 
    We use the existing TSMD technique LoCoMotif~\citep{locomotif} in~(c) and the proposed one, \locomotifdok, in~(d).
    }
    \label{fig:physical_therapy_example}
\end{figure}

Due to the unsupervised nature of TSMD, the identified motif sets might not be aligned with the user's intentions.
To demonstrate this, consider a physiotherapy use case 
where a subject performs certain exercises repeatedly and their body motions are captured by wearable sensors as time series data. %
In such a time series, 
the desired motifs are individual exercise executions grouped into motif sets according to their execution type.
Figure~\ref{fig:physical_therapy_example}a depicts an example time series 
that contains an exercise performed in three different execution types, each type repeated ten times.
In this time series, an unsupervised TSMD technique fails to discover the desired motif sets (Figure~\ref{fig:physical_therapy_example}b); instead, it mostly finds motif sets that consist of parts of exercise executions and idle time periods between them
(Figure~\ref{fig:physical_therapy_example}c). 
We call such motif sets, which are not interesting to the user, \definition{off-target} motif sets.
Off-target motif sets are also discovered in the rest of the physiotherapy dataset (Section~\ref{sec:physiotherapy}) as well as in other application domains such as electrocardiogram, seismology, and epileptic seizure monitoring (Section~\ref{sec:experimental_evaluation} and~\cite{guiding}).

To discover desired motif sets in TSMD, we leverage \definition{domain knowledge}, which is the available information about the desired motifs and can be defined by an expert without much effort in many applications. 
Domain knowledge includes properties of motifs or motif sets to be discovered, such as restrictions on and preferences for certain patterns, and does \textit{not} require manually annotating occurrences of any patterns. 
In our physiotherapy example, we have three pieces of domain knowledge: 
(1)~The subject was advised to perform the exercises ten times, %
(2)~every exercise execution starts and ends when the subject is at resting position, %
and 
(3)~each execution type has a specific range of duration.
Leveraging these three pieces of domain knowledge enables us to exclusively discover the desired motif sets in the example time series (Figure~\ref{fig:physical_therapy_example}d).

Although incorporating domain knowledge into TSMD is valuable, only a few existing TSMD techniques support it, and their support is quite limited.
To represent domain knowledge, these techniques merely consider an auxiliary time series that specifies for every time index whether a motif can start (or end) there or how preferable such a motif is~\citep{guiding,mohammad2009constrained,locomotif}.
This approach is not expressive enough to represent a wide variety of domain knowledge; for example, to define the knowledge in our running example%
. 
Some TSMD techniques have hyperparameters (such as motif length) that can be selected based on domain knowledge, but they are also not expressive enough. 
For example, a hyperparameter that specifies the motif length range usually applies to all discovered motifs, whereas the domain knowledge in our running example includes different ranges for different motif sets to be discovered.
Therefore, most forms of domain knowledge cannot be leveraged by existing TSMD techniques.

This paper first proposes a framework for incorporating domain knowledge into TSMD in a generic and highly expressive manner. 
The framework supports domain knowledge 
expressed by hard and soft constraints that respectively represent the restrictions on and preferences for the patterns to be discovered. 
Constraints can be defined differently for individual motif sets and for specific pairs of motif sets.
Defining constraints does not require any knowledge about the TSMD technique being used. 
Although constraints can easily be defined by the user, for convenience, we provide a catalogue of %
constraints that might be useful in multiple applications.

Secondly, this paper proposes an implementation of the proposed framework based on an existing TSMD technique, LoCoMotif~\citep{locomotif}, leading to \definition{\locomotifdok} (LoCoMotif with Domain Knowledge). 
We evaluate \locomotifdok based on three different experiments with different types of domain knowledge: 
a physiotherapy use case, a benchmark for TSMD, and time series from other applications. 
In general, \locomotifdok successfully leverages domain knowledge, outperforming existing TSMD techniques that support domain knowledge.

The remainder of this paper is organized as follows: 
After providing the notation (Section~\ref{sec:notation}), we describe our proposed framework for TSMD with domain knowledge (Section~\ref{sec:framework_with_domain_knowledge}) and its implementation \locomotifdok (Section~\ref{sec:locomotif_dok}). 
We then provide an overview of the related work (Section~\ref{sec:related_work}).
We evaluate \locomotifdok in a physiotherapy use case, on a TSMD benchmark, and in other applications (Section~\ref{sec:experimental_evaluation}). 
Lastly, we draw conclusions (Section~\ref{sec:conclusion}).

\section{Notation}\label{sec:notation}
A \definition{time series} $\mathbf{x}$ of length $n$ is a sequence of $n$ samples (i.e., measurements):\ 
${\mathbf{x} = [x_1, \ldots, x_n]}$ %
where $x_i \in \mathbb{R}$ for univariate time series and $x_i \in \mathbb{R}^D$ for $D$-dimensional multivariate time series for $i=1,\ldots,n$. 

A \definition{(time) segment} $\beta = [b:e] = [b, b+1, \ldots, e]$ is an interval of time indices with length $|\beta| = e-b+1$ %
for $1 \leq b \leq e \leq n$.
The \definition{subsequence} of $\mathbf{x}$ that corresponds to $\beta$ is $\mathbf{x}_\beta = [x_b, x_{b+1}, \ldots, x_e]$.
To check whether a segment $\beta$ overlaps with another segment $\beta'$ excessively or not, we define that $\beta$ is \definition{$\nu$-coincident} to $\beta'$ if \mbox{$|\beta \cap \beta'| > \nu \cdot |\beta'|$} where $\nu$ is the \definition{overlap} parameter.

A \definition{motif} in $\mathbf{x}$ is a subsequence $\mathbf{x}_\beta$ (which corresponds to the segment $\beta$) that contains a pattern that occurs multiple times in $\mathbf{x}$. 
Occurrences of the same pattern constitute a \definition{motif set}%
:\ $\mset = \{\beta_1, \ldots, \beta_k\}$%
, where $|\mset| = k$ is the \definition{cardinality} of $\mset$%
. 
The \definition{time coverage} of $\mset$ is defined as the overall time duration covered by the motifs: $\text{cov}(\mset) = \left| \bigcup_{\beta \in \mset} \beta \right|$.

\section{Proposed Framework for Motif Discovery with Domain Knowledge}\label{sec:framework_with_domain_knowledge}

We introduce a TSMD framework that allows the user to specify domain knowledge 
as constraints on the motif sets to be discovered. 
By only finding motif sets that comply to the specified constraints, we are more likely to discover the desired patterns. 
We formulate this task as a constrained optimization problem, which is explained next.

\subsection{Constrained Optimization Problem}\label{sec:constrained_optimization}

The constrained optimization problem has the decision variables $\mset_1, \ldots, \mset_\kappa$ that represent the $\kappa$ motif sets to be discovered.
Constraints are defined separately for every motif set $\mset_i$ and for every pair of distinct motif sets $(\mset_i, \mset_j)$ where \mbox{$i,j \in [1:\kappa]$ with $i \neq j$}.

\begin{figure}
    \centering
    % \includesvg[width=.85\textwidth]{figures/LoCoMotif_DoK_constraints_v2.svg}%
    \includegraphics[width=.85\textwidth]{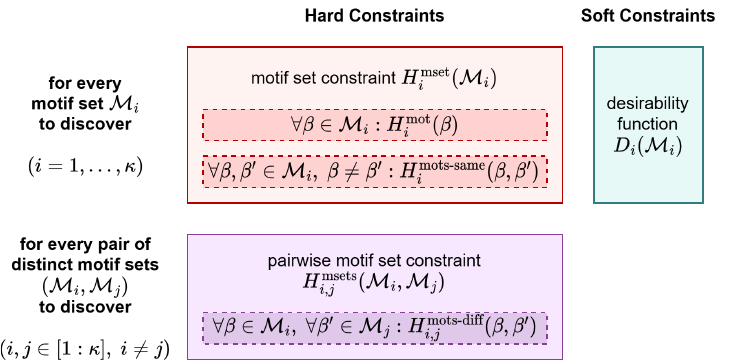}
    \caption{%
    Taxonomy of constraints.
    Dashed rectangles indicate special cases of constraints.
    }
    \label{fig:taxonomy_of_constraints}
\end{figure}

Constraints can be of two types:
hard constraints (requirements) and soft constraints (preferences).
Hard constraints must be satisfied by the discovered motif sets, and are represented by predicates %
that evaluate to True when 
satisfied %
and False otherwise. 
Figure~\ref{fig:taxonomy_of_constraints} shows a taxonomy of the supported constraints. 
We support two types of hard constraints: 
\definition{hard motif set constraints} $\Hmset_i(\mset_i)$ ($i\in [1:\kappa]$) that are defined for individual motif sets, 
and \definition{hard pairwise motif set constraints} $\Hmsets_{i,j}(\mset_i, \mset_j)$ (where $i,j \in [1:\kappa]$ with $i \neq j$) that are defined for pairs of distinct motif sets. 
A soft constraint is implemented through a \definition{desirability function} $\desir_i(\mset_i)$ ($i\in [1:\kappa]$) that determines a motif set's desirability by a scalar ranging from 0 (completely undesirable) to 1 (most desirable).%
\footnote{%
We do \textit{not} consider 
desirability functions defined on pairs of motif sets
because it would not be straightforward to take them into account in the objective function of the optimization problem.}
To discover the \textit{best} motif sets, we consider a nonnegative \definition{quality measure} $\varphi(\mset)$ that evaluates a motif set $\mset$ in an unsupervised manner, 
e.g., based on the similarities between motifs %
as existing TSMD techniques do~\citep{locomotif,motiflets}.

These concepts lead us to the constrained optimization problem
\begin{equation}\label{eq:constrained_optimization}
\begin{aligned}
    \max_{ \mset_1, \mset_2, \dots, \mset_{\kappa}} \quad & \sum_{i=1}^{\kappa}{\Dmset_i(\mset_i) \cdot \varphi(\mset_i)} \\    
    \textrm{s.t.} \quad     &  \Hmset_i(\mset_i) && \forall i \in [1:\kappa] \\ 
                            &  \Hmsets_{i,j}(\mset_i, \mset_j) && \forall i, j \in [1:\kappa],\; i \neq j \\
\end{aligned}
\end{equation}
whose solution $\mset_1^*, \ldots, \mset_\kappa^*$ consists of the discovered motif sets.

The objective function is the sum of the qualities $\varphi(\mset_i)$ of the motif sets weighted by the desirability functions $\Dmset_i(\mset_i)$ for $i=1,\ldots,\kappa$.
Therefore, the constrained optimization problem discovers the $\kappa$ motif sets that satisfy the hard constraints and that lead to the highest total quality weighted by the soft constraints. 

The size of the search space of the optimization problem is astronomical: 
Without any constraints, 
it is%
\footnote{
There are $n_\text{segments} \triangleq n + (n-1) + (n-2) + \ldots + 2 + 1 = \frac{n^2+n}{2}$ distinct time segments. %
Any set of distinct segments with a minimum cardinality of two constitute a motif set; hence, there are $n_\text{motif sets} \triangleq 2^{n_\text{segments}} - n_\text{segments} - 1$ distinct motif sets. 
The search space contains any ordered list of $\kappa$ motif sets; hence, its size is 
$(n_\text{motif sets})^\kappa = \left( 2^{\left(n^2+n\right)/2} - \frac{n^2+n}{2} - 1 \right)^\kappa 
\in O\!\left(2^{\kappa n^2}\right)$.
} 
is $O\!\left(2^{\kappa n^2}\right)$
and becomes a 6021-digit number when searching for merely $\kappa=2$ motif sets in a relatively short time series of length $n=100$. 
Hence, it is not possible to solve this problem using a brute force approach that evaluates every possible $\kappa$ motif sets in terms of their admissibility, desirability, and quality.
Therefore, we provide the \locomotifdok algorithm that solves this problem efficiently using a greedy strategy in Section~\ref{sec:locomotif_dok}. 

Before describing \locomotifdok, we 
provide special cases of the aforementioned constraints 
and a catalogue of constraints that might be useful in different applications.

\subsection{Special Cases of Constraints}\label{sec:special_cases_of_constraints}

While all the constraints in the optimization problem are defined on (pairs of) motif sets, we consider certain special cases of hard constraints where constraints are defined on (pairs of) motifs. 
These special cases are easier to define in certain scenarios and can be converted to soft constraints.

We consider two special cases of a motif set constraint $\hmset(\mset)$, as shown in the dashed red rectangles in Figure~\ref{fig:taxonomy_of_constraints}: 
The first one requires all motifs $\beta$ in the motif set to satisfy a predicate $\hmot(\beta)$. 
The second requires all pairs of distinct motifs $(\beta,\beta')$ in the motif set to satisfy a predicate $\hmotssame(\beta,\beta')$.
For a pairwise motif set constraint $\hmsets(\mset, \mset')$, we consider one special case (the dashed purple rectangle) where 
every pair of motifs, with one motif from each of the two motif sets, must satisfy a predicate $\hmotsdiff(\beta, \beta')$.
When applicable, the predicates in these special cases are easier to define than $\hmset(\mset)$. 
For example, one can simply define ${\hmot(\beta) \;:\; 50 \leq |\beta| \leq 100}$ to restrict the motif length.

In principle, every hard constraint that is represented by the predicate $\hmot(\beta)$ or $\hmotssame(\beta, \beta')$ can be converted into a desirability function on the corresponding motif set $\mset$ by considering the proportion of the motifs %
or the pairs of distinct motifs %
that satisfy these constraints, as follows (where $\llbracket \cdot \rrbracket$ denote the Iverson brackets): 
\begin{align}
    \dmset(\mset) &= \frac{1}{|\mset|} \sum_{\beta \in \mset} \llbracket \hmot(\beta) \rrbracket \label{eq:penalty_for_hmot} \\
    \dmset(\mset) &= \frac{1}{|\mset|^2 - |\mset|} \sum_{\substack{ \beta, \beta' \in \mset \\ \beta \neq \beta'}} \llbracket \hmotssame(\beta, \beta') \rrbracket \label{eq:penalty_for_hmotssame}
\end{align}

\subsection{Catalogue of Constraints}\label{sec:example_constraints}

The user can represent domain knowledge by defining any function in the form of \mbox{$\hmset$, $\hmot$, $\hmotssame$, $\hmotsdiff$, or $\desir$}. 
However, for convenience, we provide a catalogue of readily-implemented constraints (in Table~\ref{tab:example_constraints}) that are leveraged in existing TSMD techniques or in the remainder of this paper. 
It is important to note that desirability functions ($\desir$) can be defined differently from what we provide.

\begin{table}[htbp]
    \centering
    \scriptsize
    \begin{tabular}{L{0.1cm}L{1.6cm}L{4.3cm}L{5.9cm}}
        \hline
        \multicolumn{2}{L{2cm}}{\textbf{Type of Domain Knowledge}} & \textbf{Hard Constraint} & \textbf{Soft Constraint \newline (Desirability Function)} \newline ($\varrho \in (0,1)$ is specified by the user.) %
        \tabularnewline \hline \hline
        1 & Min.\ cardinality $k_{\min}$ &  $\hmset(\mset) \;:\; |\mset| \geq k_{\min}$ & \mbox{$\dmset(\mset) = \begin{cases} \frac{|\mset|}{k_{\min}} & \text{if} \ |\mset| < k_{\min} \\ 1 & \text{otherwise} \end{cases}$} \tabularnewline \hline
        2 & Max.\ cardinality $k_{\max}$ & 
            $\hmset(\mset) \;:\; |\mset| \leq k_\mathrm{max}$ & \mbox{$\dmset(\mset) = \begin{cases}
                \varrho^{|\mset| - k_{\max}} & \text{if $|\mset| > k_{\max}$} \\ 
                1 & \text{otherwise}
            \end{cases}$} 
            \tabularnewline \hline
        3 & Min.\ time coverage $c_{\min}$ & 
            $\hmset(\mset) \;:\; \text{cov}(\mset) \geq c_{\min}$ 
            & 
            \mbox{$\dmset(\mset) = \begin{cases}
                \frac{\text{cov}(\mset)}{c_{\min}} & \text{if $\text{cov}(\mset) < c_{\min}$} \\ 
                1 & \text{otherwise}
            \end{cases}$} \tabularnewline \hline
        4 & Max.\ time coverage $c_{\max}$ & 
            $\hmset(\mset) \;:\; \text{cov}(\mset) \leq c_{\max}$ %
            & 
            \mbox{$\dmset(\mset) = \begin{cases}
                \varrho^{\text{cov}(\mset)-c_{\max}} & \text{if $\text{cov}(\mset) > c_{\max}$} \\ 
                1 & \text{otherwise}
            \end{cases}$} \tabularnewline \hline

        5 & Length range $(l_{\min}, l_{\max})$ & $\hmot(\beta) \;:\; l_{\min} \leq |\beta| \leq l_{\max} $ & 
            $\dmset(\mset) = \prod_{\beta \in \mset}{r(\beta)}$ where 
            \mbox{$r(\beta) = 
            \begin{cases}
                \frac{|\beta|}{l_{\min}} & \text{if $|\beta| < l_{\min}$} \\ 
                \varrho^{|\beta|/l_{\max} - 1} & \text{if $|\beta| > l_{\max}$} \\ 
                1 & \text{otherwise}
            \end{cases}
            $}
            \tabularnewline \hline 
        6 & Min.\ standard deviation $\sigma_{\min}$ & $\hmot(\beta) \;:\;  \text{std}{( \mathbf{x}_\beta )} \geq \sigma_{\min}$ where $\text{std}(\mathbf{x}_\beta)$ is the standard deviation of the subsequence $\mathbf{x}_\beta$ & 
            $\dmset(\mset) = \prod_{\beta \in \mset}{r'(\beta)}$ where 
            \mbox{$r'(\beta) = 
            \begin{cases}
                \frac{\text{std}{( \mathbf{x}_\beta )}}{\sigma_{\min}} & \text{if $\text{std}{( \mathbf{x}_\beta )} < \sigma_{\min}$} \\ 
                1 & \text{otherwise}
            \end{cases}
            $}
            \tabularnewline \hline
        7 & Beginning mask $\mathbf{m}^{\text{start}}$
        & $\hmot([b:e]) \;:\; m^{\text{start}}_b$ \newline where $\mathbf{m}^{\text{start}} \in \{ 0, 1 \}^n$ %
            & $\dmset(\mset) = \prod_{[b:e] \in \mset}{m^{\text{start}}_b}$ \newline where $\mathbf{m}^{\text{start}} \in [ 0, 1 ]^n$ %
            \tabularnewline \hline 
        8 & End mask $\mathbf{m}^{\text{end}}$
        & $\hmot([b:e]) \;:\; m^{\text{end}}_e$ \newline where $\mathbf{m}^{\text{end}} \in \{ 0, 1 \}^n$ %
            & $\dmset(\mset) = \prod_{[b:e] \in \mset}{m^{\text{end}}_e}$ \newline where $\mathbf{m}^{\text{end}} \in [ 0, 1 ]^n$ %
            \tabularnewline \hline 
        9 & Overlap within a motif set & $\hmotssame(\beta, \beta') \;:\; \beta \text{ is not } \nu\text{-coincident to } \beta'$ &  Eq.~(\ref{eq:penalty_for_hmotssame}) \tabularnewline \hline
        10 & Overlap between motif sets & $\hmotsdiff(\beta, \beta') \;:\; \beta \text{ is not } \nu\text{-coincident to } \beta'$ & N/A \tabularnewline \hline
    \end{tabular}
    \caption{Catalogue of constraints.} 
    \label{tab:example_constraints}
\end{table}

Types~1--4 of domain knowledge restrict two characteristics of a motif set: cardinality and time coverage. 
The corresponding hard constraints set lower- or upper-bounds on these characteristics. 
In the soft implementation, desirability is set to 1 when the characteristic is within the bounds, and is gradually decreased beyond the bounds. 
The decrease is 
linear beyond the lower bound (where it reaches 0 when the characteristic becomes 0) and 
exponential beyond the upper bound with a decay constant $\varrho \in (0,1)$ provided by the user.%
\footnote{For the lower bound, we use a linear penalization strategy to avoid requiring a parameter for penalization.} 
For example, for a given cardinality range of $(k_{\min}, k_{\max}) = (5, 7)$, motif sets of cardinalites 2 and 7 have desirabilities $2/5$ and $\varrho^2$, respectively.

Types~5--8 of domain knowledge impose constraints on individual motifs according to their length, standard deviation of the measurements in them, and whether their beginning and end time points are allowed or not (determined by the \emph{mask}s). %
Their hard implementations simply require defining $\hmot$ according to these properties. 
Their soft implementations, on the other hand, deserve some explanation: 
For types~5--6, we set a desirability $r(\beta)$ for every motif $\beta$ in the motif set (similarly to the desirability in types~1--4), and multiply these desirabilities to obtain the overall desirability $\Dmset$ of the motif set. 
For types~7--8, we rely on user-provided \emph{soft mask}s that represent the preference (between 0 and 1) for a motif starting or ending at every time index. 
We then compute the desirability of a motif set by multiplying the preferences for all motifs.

Lastly, types~9~and~10 involve preventing overlaps between motifs that are in the same motif set (type~9) or in different motif sets (type~10), where 
the extent of overlap is controlled by the parameter $\nu\in[0,1]$. 
While $\nu=0$ strictly prevents overlaps, $\nu=1$ completely removes the overlap constraint so that any overlaps are allowed. 

Our physiotherapy example (Figure~\ref{fig:physical_therapy_example}(d)) uses 
$\kappa = 3$, soft constraints of types 1, 2 
and hard constraints of types 5, 7, 8, 9, 10. 
Among them, the constraints of types~1~and~2 are applied with different parameters on the three motif sets to be discovered.%

\section{Implementation of the Proposed Framework: \locomotifdok}\label{sec:locomotif_dok}

As an implementation of the proposed framework, we present the \locomotifdok algorithm. 
We first provide an overview of \locomotifdok and then explain individual mechanisms of it. 

\subsection{Overview and Motivation}\label{sec:locomotif_dok_overview}

\locomotifdok is a TSMD technique that discovers motif sets that satisfy hard constraints by taking into account soft constraints. %
\locomotifdok performs this by solving the constrained optimization problem in Equation~\ref{eq:constrained_optimization} in a tractable manner. 
Hence, \locomotifdok supports all types of constraints and desirability functions that are considered in the framework. 
In our implementation\footnote{The source code is available at \url{https://github.com/ML-KULeuven/locomotif_dok}.}, those constraints and desirability functions can simply be selected from the catalogue (Table~\ref{tab:example_constraints}) or defined by the user as Python functions.

We base \locomotifdok on the existing TSMD technique LoCoMotif~\citep{locomotif} for two reasons. 
First, LoCoMotif %
has useful capabilities: It can discover multiple motif sets of different cardinalities where motif length can differ within and among motif sets; it supports multivariate time series; and it can optionally apply time warping to compensate for nonlinear variations in time. 
Second, the LoCoMotif algorithm
efficiently obtains a diverse set of so-called candidate motif sets, evaluates them, and selects the highest-quality ones as the discovered motif sets. 
This process is well-suited for integrating domain knowledge 
by filtering out inadmissible motif sets and scaling the quality of the remaining ones by their desirability.

\begin{figure}
    \centering
    %
    % \includesvg[width=\textwidth]{figures/LoCoMotif_DoK_flowchart_v5.svg}%
    \includegraphics[width=\textwidth]{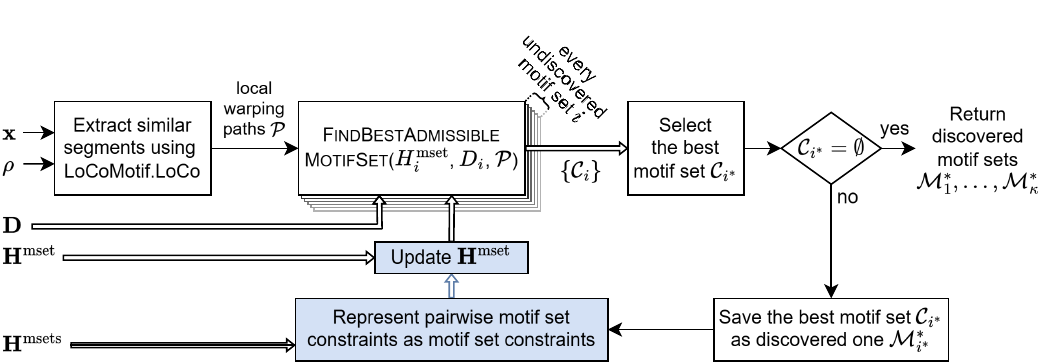}
    \caption{%
    Flowchart of \locomotifdok. 
    }
    \label{fig:locomotif_dok_flowchart}
\end{figure}

We provide the pseudocode of \locomotifdok in Algorithm~\ref{algo:locomotif_dok} and its flowchart in Figure~\ref{fig:locomotif_dok_flowchart}. %
\locomotifdok first 
extracts time segments of $\mathbf{x}$ that are similar to each other using the Local Concurrences (LoCo) method of LoCoMotif (line~1)
so that candidate motif sets can later be generated efficiently. 
Then, \locomotifdok discovers $\kappa$ motif sets greedily (lines~4--14), where $\kappa$ is specified by the user: 
For each motif set $i$ to discover ($i=1,\ldots,\kappa$), 
the best admissible motif set $\cmset_i$ is found by the procedure \textsc{FindBestAdmissibleMotifSet} by taking into account the domain knowledge provided for the $i$-th motif set. 
Among these $\kappa$ motif sets, the best one ($\cmset_{i^*}$) is saved as the $i^*$-th discovered motif set $\mset_{i^*}^*$. 
This process is repeated for the undiscovered motif sets until all $\kappa$ are discovered or no more admissible motif sets can be found.

\begin{algorithm}[!htbp]
\caption{\locomotifdok}\label{algo:locomotif_dok}
\SetAlgoLined
\DontPrintSemicolon
\SetKwInOut{KwIn}{\hspace{-0.25cm}Input}
\SetKwInOut{KwOut}{\hspace{-0.25cm}Output}
\SetKwFunction{findbams}{FindBestAdmissibleMotifSet}
\SetKwFunction{filtercmset}{FilterCandidateMotifSet}
\SetKwProg{myproc}{procedure}{}{}
\SetKw{Or}{or}
\KwIn{%
    $\mathbf{x}$: time series, 
    $\rho$: strictness, 
    $\kappa$: max.\ number of motif sets to discover, \newline
    $\HMSET = [\,\Hmset_i\,]_{i=1}^\kappa$: motif set constraints, \newline
    $\HMSETS = [\,\Hmsets_{i,j}\,]_{i,j \in [1:\kappa],\; i \neq j}$: pairwise motif set constraints, \newline
    $\DMSET = [\,\Dmset_i\,]_{i=1}^\kappa$: desirability functions. 
}
\KwOut{$\{ \mset_1^*, \ldots, \mset_\kappa^* \}$: discovered motif sets.}
$\mathcal{P} \gets \text{LoCoMotif.LoCo}(\mathbf{x}, \rho)$ \tcp{Obtain the set $\mathcal{P}$ of local warping paths.} 
$\mathcal{I} \gets [1:\kappa]$ \tcp{indices of motif sets to be discovered}
$\mset_i \gets \emptyset \quad \text{for~} i=1,\ldots,\kappa$ \tcp{initialize every motif set to be discovered}
\While{$\mathcal{I} \neq \emptyset$}{
    $\cmset_i \gets \emptyset, \quad \phi_i \gets 0 \quad \text{for~} i=1,\ldots,\kappa$\;
    \For {$i \in \mathcal{I}$}{
        $\cmset_i, \phi_i \gets$ \findbams{$\Hmset_i$, 
        $\Dmset_i$, 
        $\mathcal{P}$
        }
    }
    $i^* \gets \argmax{[\phi_1,\ldots,\phi_\kappa]}$ \tcp{index of the best discovered motif set}
    \If{$\cmset_{i^*} = \emptyset$}{Exit the loop. \tcp*[h]{no admissible motif sets found}}
    $\mset_{i^*}^* \gets \cmset_{i^*}$ \tcp{Save $\cmset_{i^*}$ as the ${i^*}^\text{th}$ discovered motif set.}
    $\mathcal{I} \gets \mathcal{I} \setminus \{ i^* \}$ \tcp{Remove the index of the discovered motif set.}
    \For(\tcp*[h]{every undiscovered motif set $j$}){$j \in \mathcal{I}$}{
        $\Hmset_j(\cmset) \gets \Hmset_j(\cmset) \land \Hmsets_{j,i^*}(\cmset, \mset_{i^*}^*) \land \Hmsets_{i^*,j}(\mset_{i^*}^*, \cmset)$ 
        \tcp{Represent pairwise motif set constraints as motif set constraints.
        }
    }
}
\Return{%
    $\{ \mset_1^*, \ldots, \mset_\kappa^* \}$
    }

\vspace{1mm} \hrule \vspace{1mm}

\myproc{\findbams{%
        $\Hmset$,
        $\Dmset$, 
        $\mathcal{P}$
    }}{
    $\{ \Hmot, \Hmotrepr, \Hmotssame, k_\mathrm{max}^\text{discard}, \Hmset{}^\text{-others} \} \gets \Hmset$ \tcp{\mbox{Categorize $\Hmset$.}}
    $(\cmset^*, \phi^*) \ \gets (\emptyset, 0)$ \tcp{initialize the best motif set with its quality}
    \For {$b$ \In $[1:n]$}{
        \For {$e$ \In $[b:n]$}{
            $\alpha \gets [b : e]$ \tcp{representative segment} 
            \If{$\Hmot(\alpha) \,\land\, \Hmotrepr(\alpha)$ 
            }{%
                $\cmset \gets \text{LoCoMotif.generateCandidateMotifSet}(\alpha, \mathcal{P})$ \;
                $\cmset \gets$ \filtercmset{$\cmset$, $\Hmot$, $\Hmotssame$, $k_{\max}^\text{discard}$} \; 
                \If{$|\cmset| \geq 2 \,\land\, \Hmset{}^\text{-others}(\cmset)$ 
                }{
                    $\phi \gets \varphi(\cmset) \cdot \Dmset(\cmset)$ \tcp{quality of $\cmset$ weighted by its desirability}
                    \lIf{$\phi > \phi^*$}{$(\cmset^*, \phi^*) \gets (\cmset, \phi)$}
                }
            }
        }
    }
    \Return $\cmset^*$, $\varphi^*$
}

\vspace{1mm} \hrule \vspace{1mm}

\myproc{\filtercmset{%
        $\cmset$, $\Hmot$, $\Hmotssame$, $k_{\max}^\text{discard}$
    }}{
        $\cmset_\text{new} \gets \emptyset$ \;
        \For{$\beta$ \In $\cmset$}{
            \If{%
                $|\cmset_\text{new}|<k_{\max,\,i}^\text{discard} \,\land\, \Hmot(\beta) \,\land\, \Hmotssame(\beta, \beta')\;\; \forall \beta' \in \cmset_\text{new}$
                }{%
                $\cmset_\text{new} \gets \cmset_\text{new} \cup \{ \beta \}$
            }
        }
        \Return $\cmset_\text{new}$
}

\end{algorithm} 

In the next subsections, we explain the mechanisms of 
generating a candidate motif set, %
finding the single best admissible motif set, %
and
discovering multiple motif sets%
.

\subsection{Generating A Candidate Motif Set}\label{sec:preprocessing_and_generating_a_cmset}

\begin{figure}
    \begin{subfigure}[t]{0.44\linewidth}
        \includegraphics[width=\linewidth]{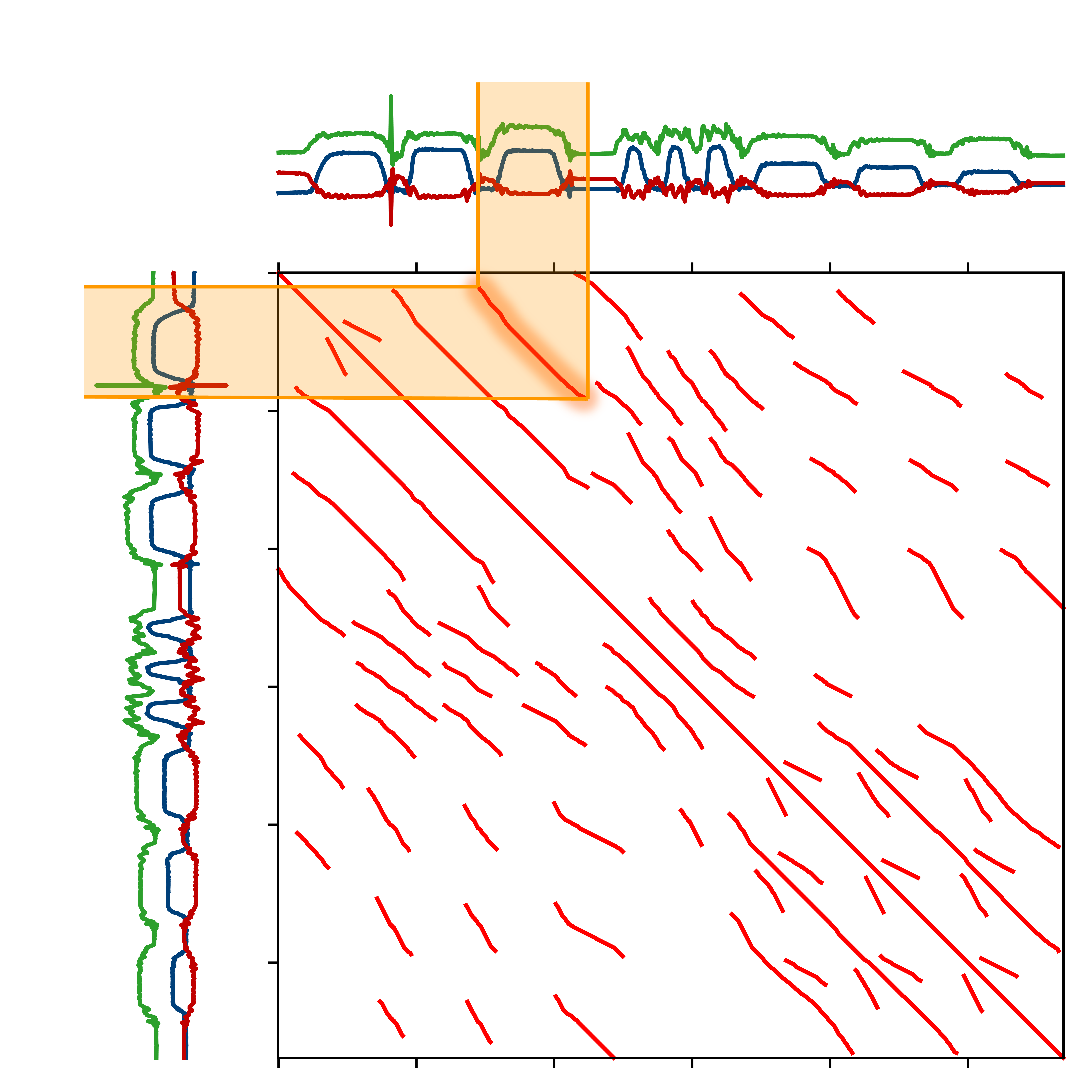}
        \label{fig:local_warping_paths_1}
    \end{subfigure}
    \hfill
    \begin{subfigure}[t]{0.44\linewidth}
        \includegraphics[width=\linewidth]{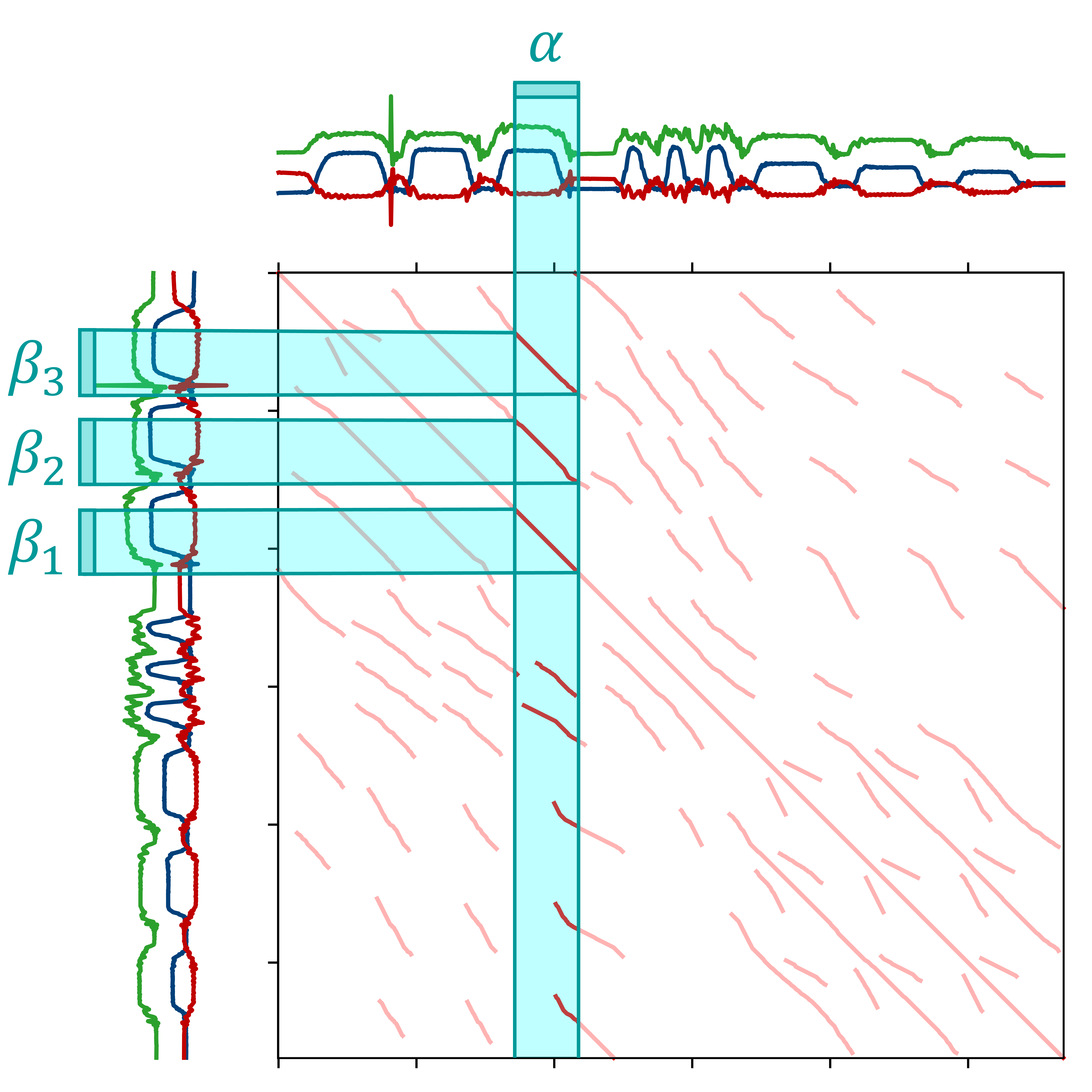}
        \label{fig:local_warping_paths_2}
    \end{subfigure}
    \caption{\textit{Left:}~LoCo relates similar time segments of $\mathbf{x}$ with each other by local warping paths $\mathcal{P}$, where one of the paths is highlighted. \textit{Right:}~From a given representative segment $\alpha$, a candidate motif set $\cmset = \{\beta_1, \beta_2, \beta_3\}$ is obtained efficiently by identifying segments that are similar to $\alpha$ according to $\mathcal{P}$.}
    \label{fig:local_warping_paths}
\end{figure}

\locomotifdok inherits LoCoMotif's LoCo step~\citep{locomotif} that enables obtaining candidate motif sets efficiently.
LoCo obtains a set $\mathcal{P}$ of local warping paths, each of which relates two similar time segments of $\mathbf{x}$ with each other with time warping, as shown in Figure~\ref{fig:local_warping_paths}~(left). 
The strictness of assessing the similarity between segments is controlled by the hyperparameter $\rho \in [0, 1]$: 
A larger $\rho$ requires the segments to be more similar to each other to be related by a local warping path. %

To generate a candidate motif set, we use the function LoCoMotif.generateCandidateMotifSet, which efficiently obtains all the segments $\beta_1, \ldots, \beta_k$ that contain a similar pattern to a given segment $\alpha$ by processing $\mathcal{P}$ as shown in Figure~\ref{fig:local_warping_paths}~(right). 
Here, $\alpha$ is called a \definition{representative segment} and $\cmset = \{ \beta_1, \ldots, \beta_k \}$ the corresponding \definition{candidate motif set}. 
The segments in $\cmset$ are sorted from the most similar to $\alpha$ to the least similar%
\footnote{%
We measure the similarity between every motif $\beta \in \cmset$ and $\alpha$ 
by summing up the similarities on the subpath $\mathbf{q}$ of the partial warping path $\mathbf{p} \in \mathcal{P}$ that relates $\beta$ to $\alpha$:
$%
    \sum_{(i, j) \in \mathbf{q}} {s(x_i, x_j)}
$
where $s(x_i, x_j) \triangleq \exp{(-\|x_i-x_j\|^{2})}$ is the similarity between $\mathbf{x}_i$ and $\mathbf{x}_j$. %
This is consistent with the way the similarity between subsequences is calculated to evaluate the quality of a candidate motif set in LoCoMotif(-DoK).%
}%
; hence, $\beta_1=\alpha$.

\subsection{Finding the Best Admissible Motif Set}\label{sec:finding_the_best_admissible_motif_set}

Given a hard motif set constraint $\Hmset$ and desirability function $\Dmset$,
the procedure \textsc{FindBestAdmissibleMotifSet} (lines~16--28 in Algorithm~\ref{algo:locomotif_dok}) finds the best admissible motif set in three steps by
(1)~generating candidate motif sets, 
(2)~filtering them (or individual motifs in them) according to 
the constraints, %
and
(3)~evaluating the quality of the remaining candidate motif sets by taking $\Dmset$ into account to eventually select the best candidate motif set.%

To be able to check the constraints at the correct stages in the algorithm, we first separate the provided hard constraints $\Hmset$ into five categories (line~17):
$\Hmot$ for constraints that are applied on every motif, 
the newly-introduced $\Hmotrepr$ for constraints that are applied on every representative motif,
$\Hmotssame$ for constraints that are applied on every pair of distinct motifs within the motif set, 
$k_{\max}^\text{discard}$ for the cardinality limit of a motif set (beyond which motifs will be removed).%
\footnote{The constraint $k_{\max}^\text{discard}$ is different from the maximum cardinality constraint $k_{\max}$ (given in Table~\ref{tab:example_constraints}), which discards the entire candidate motif set if it contains more than $k_{\max}$ motifs instead of removing excessive motifs in it.}
We introduce $\Hmotrepr$ (which is not present in the framework) here for two reasons: 
First, the user should be able to apply certain constraints only to representative motifs instead of all motifs because every motif in a motif set is already similar to the representative.
Second, we would like to support the functionality in the existing TSMD techniques (which are discussed in Section~\ref{sec:related_work}) where certain constraints are applied only on the representatives.

Then, we generate candidate motif sets only from the \textit{admissible} representative segments using LoCoMotif.generateCandidateMotifSet function. 
We check the admissibility of a representative segment $\alpha$ not only according to $\hmotrepr$ but also $\hmot$ (line~22) because the representative itself is one of the motifs in the candidate motif set and all motifs must satisfy $\hmot$.
This process results in a search space of at most $O(n^2)$ candidate motif sets, which is much smaller than that of the framework ($O(2^{(n^2)})$ for a single motif set).
Despite its manageable size, 
the reduced search space is sufficiently diverse for discovering motif sets because it includes (approximate) occurrences (if any) of every subsequence $\mathbf{x}_\alpha$ as a candidate motif set.

Next, we process every candidate motif set $\cmset$ according to the constraints (lines~24--27). 
First, we filter out individual motifs in $\cmset$ to satisfy the constraints $\Hmot$, $\Hmotssame$, and $k_{\max}^\text{discard}$, if possible.%
\footnote{We need this filtering operation because a subset of an inadmissible candidate motif set might be admissible; however, such a subset is not included in the search space because every candidate motif set contains \textit{all} occurrences of a pattern.}
The filtering operation is implemented greedily by the procedure \textsc{FilterCandidateMotifSet}, %
which starts with a new, empty candidate motif set %
and adds motifs to it one by one as long as the constraints are satisfied.
This operation adheres to the order of motifs in $\cmset$ because they are already sorted from the most relevant to the least relevant. 
Subsequently, if the (possibly filtered) candidate motif set $\cmset$ contains at least two motifs and is admissible according to $\Hmset{}^\text{-others}$, 
we evaluate its quality using the \definition{fitness} metric $\varphi(\cmset) \in [0,1]$, which considers the similarities of the motifs to the representative as well as the fraction of the time range the motifs cover, as in LoCoMotif. %
To take into account soft constraints, we scale the fitness by the desirability $\Dmset(\cmset)$. %

As a result of this process, the procedure \textsc{FindBestAdmissibleMotifSet} returns the admissible candidate motif set that has the highest fitness times desirability.
If there are no admissible motif sets found, $\emptyset$ is returned.

\subsection{Discovering Multiple Admissible Motif Sets}\label{sec:discovering_multiple_motif_sets}

\locomotifdok discovers multiple motif sets greedily:
Instead of finding $\kappa$ globally optimal motif sets%
, 
it identifies the best admissible motif set at a time by the procedure \textsc{FindBestAdmissibleMotifSet} %
and repeats this process until 
$\kappa$ motif sets are discovered or no more can be found (lines~4--14 in Algorithm~\ref{algo:locomotif_dok}).
While this iterative approach is analogous to LoCoMotif, \locomotifdok needs to ensure that the discovered motif sets satisfy constraints, which is not trivial because constraints 
might be defined differently for every (pair of) motif set(s) to be discovered.

Since the constraints might differ across the motif sets $\mset_1, \ldots, \mset_\kappa$, in every iteration of \locomotifdok, the \textsc{FindBestAdmissibleMotifSet} procedure is called %
separately for every motif set that hasn't been discovered yet%
. 
Among these `best' motif sets, the one with the highest fitness scaled by desirability is selected as the discovered motif set ($\mset_{i^*}^*$).%
\footnote{%
If the same constraints are defined for every (pair of) motif set(s), then there is no need to call the \textsc{FindBestAdmissibleMotifSet} procedure multiple times because their result would be the same. 
}
If there are no motif sets found or all motif sets have been discovered, then the iterative process is terminated. 
Hence, the \locomotifdok algorithm can discover less than $\kappa$ motif sets (in which case some $\mset_i^*$ are $\emptyset$).

The \textsc{FindBestAdmissibleMotifSet} procedure checks the constraints of types $\Hmset$ and $\Dmset$ (which are defined for individual motif sets) and not $\Hmsets$ (which are defined for pairs of motif sets). 
This is because every constraint defined for a pair of motif sets can be represented as a constraint defined for a single motif set if one of the two motif sets is known (i.e., discovered). 
Given this fact, once \locomotifdok discovers a motif set ($\mset_{i^*}^*$), for every undiscovered motif set $j$, it changes the representation of the constraints $\Hmsets_{j,i^*}$ and $\Hmsets_{i^*,j}$ (which involve $\mset_{i^*}^*$) into $\Hmset_j$ by substituting the corresponding free variables by $\mset_{i^*}^*$. 
\locomotifdok then combines the constraints whose representations have changed, with the existing constraints $\Hmset_j$ using the `and' operation in line~14, where $\cmset$ is an independent variable that represents any motif set. 
This process is represented by the parts highlighted in blue in Figure~\ref{fig:locomotif_dok_flowchart}.

\section{Related Work}\label{sec:related_work}

Most existing TSMD techniques do not support domain knowledge. 
Although they have hyperparameters, some of which can be determined using domain knowledge, they are never sufficient to represent domain knowledge in general. 
This is because hyperparameters are limited in expressiveness, as they only capture predetermined properties such as motif length or the cardinality of motif sets~\citep{locomotif,motiflets}. %

To our knowledge, there are only three existing TSMD techniques that support domain knowledge, but they are all quite limited in expressiveness%
. 
They merely utilize a mask $\mathbf{m}$, which is an auxiliary time series that specifies the desirability (between $0$ and $1$) or permissibility (either $0$ or $1$) of motifs at each time index.%
\footnote{The mask $\mathbf{m}$ is referred to as an annotation vector in~\cite{guiding} and a constraint in~\cite{mohammad2009constrained}.} 
The mask $\mathbf{m}$ can be provided by a domain expert or computed from the input time series $\mathbf{x}$ (or from an auxiliary one) using a simple method that leverages domain knowledge. 
We explain these existing techniques below%
.

\cite{guiding} proposed the technique Matrix Profile V (MPV) that applies $\textbf{m}$ to the starting time index of the \textit{representative} motif only (which is called the query). 
We can implement this in \locomotifdok by defining 
the hard constraint $\hmotrepr([b:e]) \;:\; m_b$ %
for a binary $\mathbf{m}$ 
or
the desirability function $\dmset(\{ [b_1 : e_1], \ldots, [b_k : e_k] \}) = m_{b_1}$ for a real-valued $\mathbf{m}$ 
where $b_1$ is the starting time index of the representative (i.e., the first) motif. 
We note that
MPV discovers only a single motif set with cardinality two, which can be achieved in \locomotifdok by setting $\kappa=1$ and $k_\mathrm{max}^\text{discard}=2$.

The TSMD algorithms proposed by~\cite{mohammad2009constrained} probabilistically sample the representative segments %
instead of processing all of them, where
$m_i$ specifies the probability of sampling (i.e., the probability of occurrence of) a motif that ends at time point $i$ 
for $i=1,\ldots,n$. 
If we consider the probability of occurrence as desirability, we can implement this in \locomotifdok by setting 
$\dmset(\{ [b_1 : e_1], \ldots, [b_k : e_k] \}) = m_{e_1}$ 
where $e_1$ is the end time index of the representative (i.e., the first) segment%
.
LoCoMotif~\citep{locomotif} supports two different binary masks, $\mathbf{m}^{\text{begin-repr}}$ and $\mathbf{m}^{\text{end-repr}}$, that are applied to the beginning and end time indices of \emph{representative} motifs only. 
This functionality can be implemented in \locomotifdok by defining the constraint
${\hmotrepr([b:e]) \;:\; m^{\text{begin-repr}}_b \land m^{\text{end-repr}}_e}$.

Yeh et~al.~\cite{matrixprofileIV} employs a mask for a different purpose than TSMD: 
Given a Boolean mask $\mathbf{m} \in \{ 0,1 \} ^n $ that annotates a time series $\mathbf{x}$ by specifying the time intervals containing interesting patterns, 
a model is learned to 
predict, for an unseen time series, a Boolean mask that specifies interesting time intervals in it.
This is considered as a semi- (or weekly) supervised approach 
because a model is learned from an annotated time series to make predictions for unseen time series.
As such, this method falls outside the scope of this paper.

\section{Experimental Evaluation}\label{sec:experimental_evaluation}

To evaluate the impact of domain knowledge on the motifs discovered by \locomotifdok, we perform three sets of experiments: 
detecting exercise executions in a physiotherapy use case, 
discovering the motifs inserted into time series in a TSMD benchmark, 
and 
identifying meaningful repeating patterns in several other applications.

\subsection{Physiotherapy Use Case}\label{sec:physiotherapy}

Here, we detail the physiotherapy use case, from which we have already provided an example time series in Figure~\ref{fig:physical_therapy_example}.
In this use case, we apply TSMD on wearable motion sensor data acquired during physiotherapy sessions to identify each exercise execution and group them by their execution type to aid a physiotherapist~\citep{yurtman_automated_2014}.
By using \locomotifdok, we can leverage much more comprehensive domain knowledge than LoCoMotif, which leads to the following Research Question~(RQ):

\begin{enumerate}[font=\bfseries, wide=0pt]
    \item[RQ1:] 
    By leveraging more comprehensive domain knowledge, can \locomotifdok outperform LoCoMotif  
    in identifying and grouping physiotherapy exercise executions using wearable sensor data?
\end{enumerate}

\subsubsection{Dataset} 
We use all the 40 time series that represent exercise sessions available in a publicly available dataset~\citep{physicaltherapydataset}. %
In every session, a subject performs a certain exercise in three different types: ten times in a correct manner, ten times too quickly, and ten times with a low range of motion, as shown in Figure~\ref{fig:physical_therapy_example}.
Time segments of all the executions are available as ground-truth (GT) information that we use for evaluation. 
The subjects' body movements are captured as time series data by wearable sensors (at 25~Hz) placed at five different positions on the body. 
We use the data acquired by the tri-axial accelerometer that moves the most in each exercise (position~4 for exercise~2 and position~2 for the remaining exercises), as done by~\cite{locomotif}. 

\subsubsection{Experimental Setup}
We consider 
three levels of domain knowledge:
\begin{itemize}
\item \textit{LoCoMotif-0}: 
    We do not leverage any domain knowledge.
    
\item \textit{LoCoMotif} (limited domain knowledge):
    All exercises need to start and end when the subject is at the resting position, which is determined simply by thresholding the standard deviation over a sliding window~\citep{locomotif}). 
    We represent this domain knowledge through the masks $\mathbf{m}^{\text{begin-repr}}$ and $\mathbf{m}^{\text{end-repr}}$ that constraint the beginning and end points of the representative motifs (as explained in Section~\ref{sec:related_work}). 
    This is the only type of domain knowledge LoCoMotif supports.

\item \textit{\locomotifdok} (comprehensive domain knowledge):
    In addition to the masks in the previous setting, we represent two more pieces of domain knowledge: 
    (1)~The subject is \textit{expected} to perform 10~executions of each type in every exercise session. 
    We represent this by soft constraints on the cardinality for every motif set to be discovered by setting $k_{\min} = k_{\max} = 10$ and $\varrho = 0.5$ (rows 1--2 in Table~\ref{tab:example_constraints}). 
    (2)~The three execution types present in every time series have different duration characteristics; for example, the second type is expected to be shorter than the other types (Figure~\ref{fig:physical_therapy_example}a). 
    To represent this information, we constrain the lengths of the representatives of the three motif sets to be discovered using three different length ranges (using hard constraint~5 in Table~\ref{tab:example_constraints}). 
    While a physiotherapist can define these ranges according to the expected durations in practice, we obtain them from the dataset according to the shortest and longest executions of the three execution types in the entire dataset. 
\end{itemize}

We use the same hyperparameter values for all three levels of domain knwoledge: 
We set 
$(l_{\min}, l_{\max}) = (2.0, 10.5)$ seconds 
according to the shortest and longest exercise executions in the dataset.
We repeat the experiments with different values for the strictness: $\rho \in \{ 0, 0.05, \ldots, 1 \}$. 
We use the default values for the remaining hyperparameters: $\nu = 0.25$ and time warping enabled. 

For each level of domain knowledge, we run the experiments in two settings: %
In the first setting, we assume that a physiotherapist is available. 
Since they can easily discard motif sets that do not correspond to an exercise type, we allow three additional motif sets to be discovered by setting $\kappa=6$, as in~\cite{locomotif}.
In this setting, \locomotifdok uses specialized length ranges for motif sets 1--3 and no additional length constraint for motif sets 4--6 (other than $l_{\min}$ and $l_{\max}$).
In the second setting, we assume that 
discovered motif sets are automatically processed by a downstream algorithm, where falsely discovered ones may pose a problem. 
Hence, we set $\kappa=3$.

We evaluate the discovered motifs with respect to the GT ones by using the evaluation method PROM~\citep{vanwesenbeeck2024evaluation} that computes three metrics: precision, recall, and F1-score.
High precision means fewer false positive motifs, whereas high recall means fewer missed motifs. 
F1-score is the harmonic mean of precision and recall, providing an overview evaluation of the discovered motifs. 
We set PROM to ignore falsely discovered motif sets, which is preferred in the first setting and has no effect in the second setting.

We note that other existing TSMD techniques that can leverage domain knowledge \textit{cannot} be applied to this use case due to the requirement of discovering multiple motif sets of cardinality greater than two.

\subsubsection{Results}

\begin{figure}[tb]
    \centering
    \begin{subfigure}[b]{\textwidth}
         \centering
         \includegraphics[width=\textwidth]{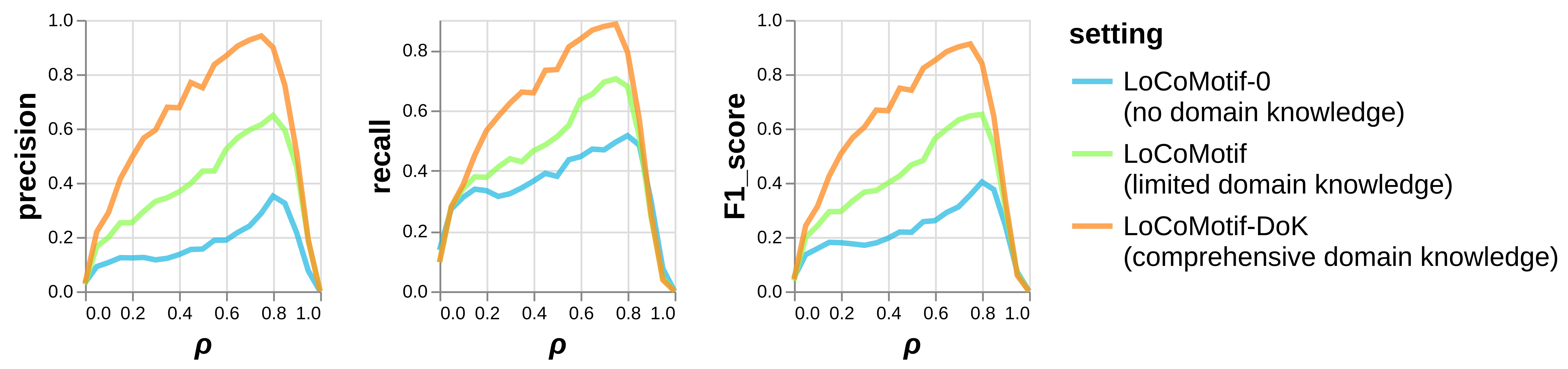}
         \caption{First setting: discovering up to $\kappa=3$ motif sets.}
         \label{fig:physiotherapy_results_a}
    \end{subfigure}
    \begin{subfigure}[b]{\textwidth}
        \vspace*{0.5cm}
        \centering
        \includegraphics[width=\textwidth]{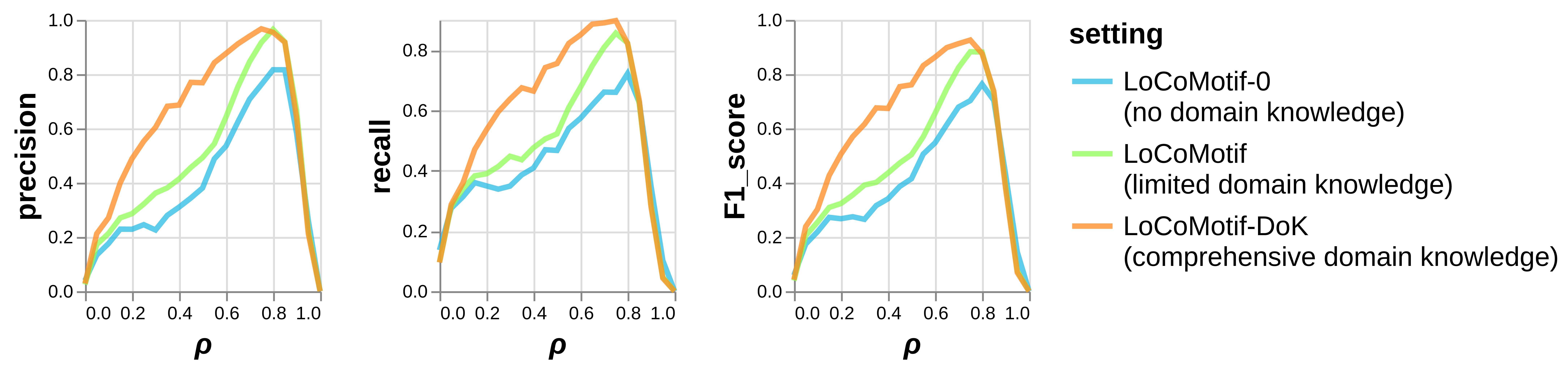}
        \caption{Second setting: discovering up to $\kappa=6$ motif sets.}
        \label{fig:physiotherapy_results_b}
    \end{subfigure}
    \caption{In the physiotherapy use case, \locomotifdok can leverage comprehensive domain knowledge and outperforms LoCoMotif that supports a much more limited form of domain knowledge in terms of different evaluation metrics for most strictness ($\rho$) values.}
    \label{fig:physiotherapy_results}
\end{figure}

Figure~\ref{fig:physiotherapy_results} provides experimental results %
for each setting, level of domain knowledge, and strictness~($\rho$). 
In both settings, \locomotifdok substantially outperforms LoCoMotif for most $\rho$ values. 
As expected, LoCoMotif-0, which doesn't leverage any domain knowledge, yields substantially worse results in all experiments, especially for $\kappa=3$. 
The main reason is the quasi-periodicity of successive exercise executions where the phase-shifted versions of the executions can be discovered as motifs, which are uninteresting repeating patterns. 
This problem is eliminated in LoCoMotif and \locomotifdok by requiring motifs to start and end when the subject is at resting position.

In terms of the peak performance (for the optimal $\rho$), \locomotifdok is dramatically better than LoCoMotif in precision, recall, and F1-score in the first setting (Figure~\ref{fig:physiotherapy_results_a}). 
For the second setting (Figure~\ref{fig:physiotherapy_results_b}), the peak performance of \locomotifdok is slightly better than LoCoMotif with a very similar precision and slightly higher recall and F1-score.
This is despite the fact that \locomotifdok is configured to discover 
up to 6 motif sets while the domain knowledge includes the presence of 3 motif sets.
We observe that the peak performance corresponds to a smaller strictness ($\rho$) when more domain knowledge is leveraged, which is expected because domain knowledge makes the algorithm stricter in a different way (through constraints) than $\rho$ (through similarities between motifs in a motif set).

To conclude,
in the physiotherapy use case, leveraging domain knowledge is essential to discover desired motifs, and additional domain knowledge can further improve the results. 
Hence, our answer to RQ1 is yes.

\subsection{TSMD Benchmark}\label{sec:benchmark}

We investigate the impact of domain knowledge %
on the quality of the discovered motif sets using a TSMD benchmark,
that consists of a diverse set of time series
for which GT motif sets are available for evaluation~\citep{vanwesenbeeck2024evaluation}. 
This leads to: 

\begin{enumerate}[font=\bfseries, wide=0pt]
    \item[RQ2:] Does incorporating additional information about the GT motif sets help in discovering motif sets that are more similar to the GT ones?
\end{enumerate}

\subsubsection{Experimental Setup}

Every time series in the benchmark consists of concatenation of time series instances from a classification dataset, where repeating classes are considered as GT motif sets.
To emulate domain knowledge in such a time series, we extract some information from each
GT motif set
and represent it by different hard and soft constraints. 
Parameters of constraints are determined specifically for every motif set in every generated time series, since the characteristics of these motif sets might differ. 

To observe the impact of constraints on the quality of the discovered motif sets, we apply LoCoMotif without domain knowledge as well as \locomotifdok with every constraint separately. 
We set the hyperparameters as $\kappa$ being the number of GT motif sets; $\nu = 0.5$; $l_{\min}$ and $l_{\max}$ being the 0.1- and 0.9-quantiles of the lengths of the instances in the validation set (of the classification dataset), as in~\cite{locomotif}.
We tune the strictness ($\rho$) and time warping hyperparameters based on a validation set~\citep{locomotif,vanwesenbeeck2024evaluation} separately for LoCoMotif and \locomotifdok with every constraint. 

We evaluate the quality of the discovered motif sets using F1-score~\citep{vanwesenbeeck2024evaluation}. 
We compare the F1-scores of each constraint with the base case through 2-D histograms as well as the number of wins, ties, and losses. 
We separate the results into two groups: hard and soft constraints. 
We further group the results according to GT motifs being fixed- or variable-length because fixed-length motifs prevent length-related constraints from being specialized for every motif set.

\subsubsection{Results}

\begin{figure}[tb]
    \centering
    \begin{subfigure}[b]{\textwidth}
         \centering
         % \includesvg[scale=0.4]{figures/histplots_g_hard_fixlen_diff.svg}
         \includegraphics[scale=0.4]{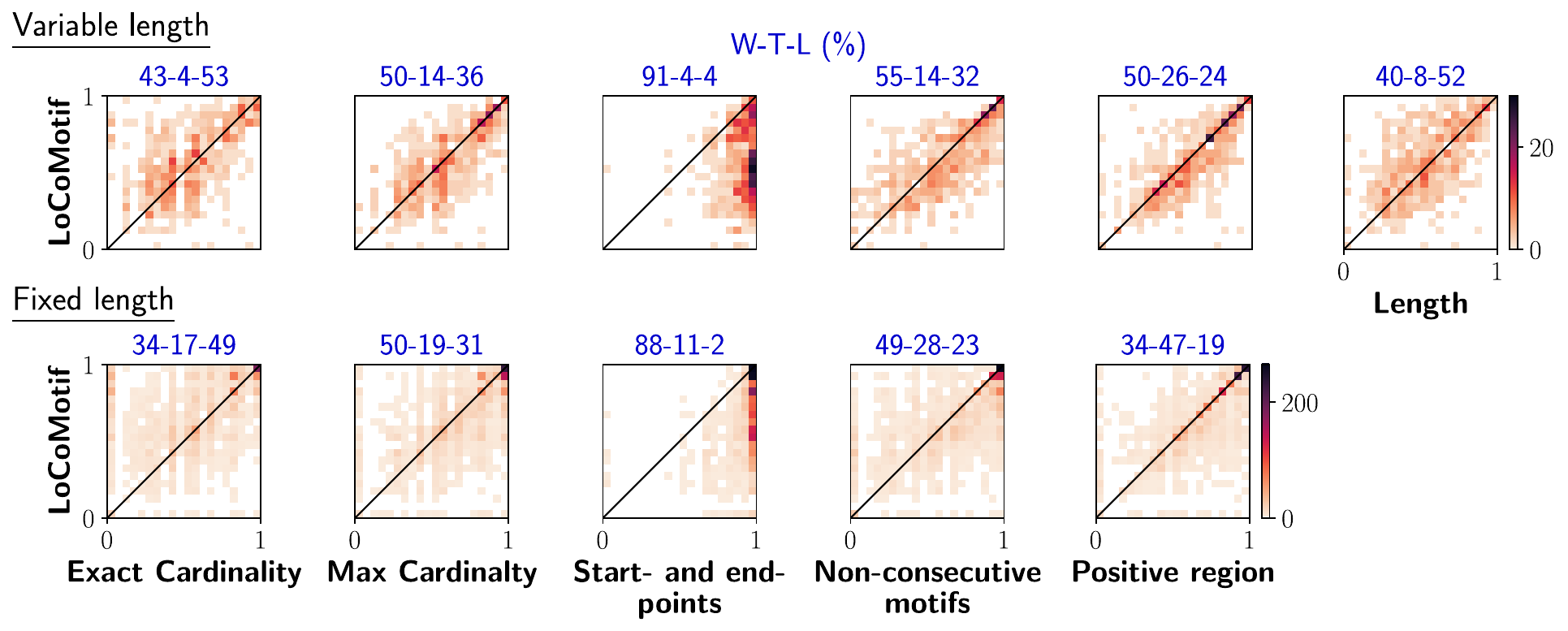}
         \caption{Hard constraints.}
         \label{fig:histplot_hard}
    \end{subfigure}
    \begin{subfigure}[b]{\textwidth}
        \centering
        % \includesvg[scale=0.4]{figures/histplots_g_soft_fixlen_diff.svg}
        \includegraphics[scale=0.4]{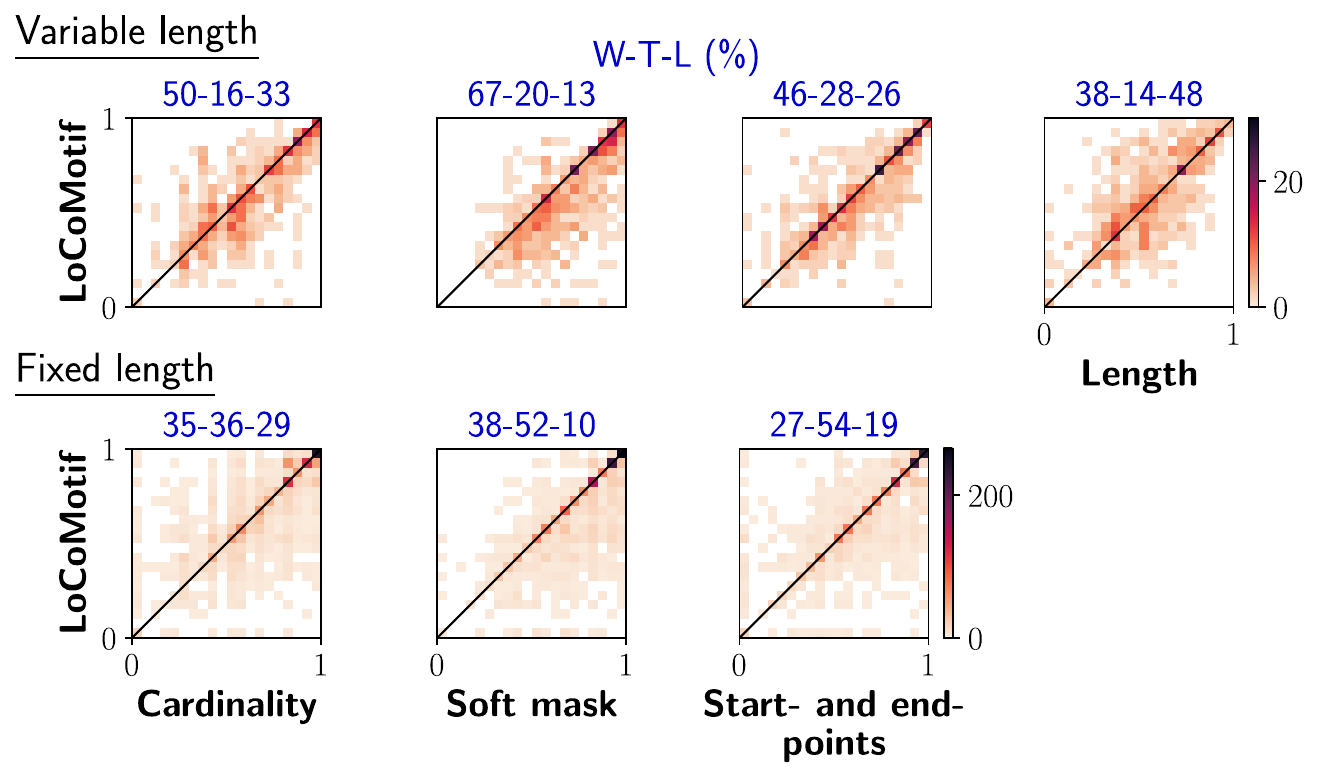}
        \caption{Soft constraints.}
        \label{fig:histplot_soft}
    \end{subfigure}
    \caption{2-D histograms that compare the F1-scores of \locomotifdok that uses every constraint with LoCoMotif without domain knowledge. The percentages of Wins-Ties-Losses are also shown.}
    \label{fig:histplots}
\end{figure}

Hard constraints are provided in Table~\ref{tab:hard_constraints_for_benchmark} and the corresponding results in Figure~\ref{fig:histplot_hard}.
The constraint Start- and End-points yields the greatest improvement, pushing the F1-scores to 1 in most cases, which is expected because the constraint defines the time segments of possible motifs with some flexibility, making the TSMD task much easier. 
The constraints Max.\ Cardinality, Non-consecutive Motifs, and Positive Region also bring an improvement. 
The Exact Cardinality constraint, on the other hand, does not help in most cases, partially because of its strictness: it discards candidate motif sets with incorrect cardinalities and may cause no motif sets to be discovered. For example, if there are five motifs in a GT motif set and one of them is too noisy, then all candidate motif sets might contain four motif sets and hence be inadmissible. 
The Length constraint also has an adverse effect on the result.

Soft constraints are listed in Table~\ref{tab:soft_constraints_for_benchmark} with their results presented in Figure~\ref{fig:histplot_soft}. 
The constraints Cardinality, Soft Mask, and Positive Region improve the F1-score. 
In particular, the soft Cardinality constraint performs much better than its hard implementation because the former does not suffer from being too strict, as it introduces a preference for certain motif sets without discarding others. 
Lastly, the Length constraint does not improve the results.

\begin{table}[tb]
    \centering
    \scriptsize
    \begin{tabular}{L{1.75cm}L{3.25cm}L{6.75cm}}
        \hline
        \textbf{Constraint Type} 
        & \textbf{Description} 
        & \textbf{Implementation} \newline (See Table~\ref{tab:example_constraints} for the catalogue.)
        \tabularnewline \hline \hline
        Exact Cardinality 
        & Require a fixed cardinality.
        & Use constraints~1--2 from the catalogue with $k_{\min}$ and $k_{\max}$ being the GT cardinality.
        \tabularnewline \hline
        Max.\ Cardinality 
        & Limit the cardinality.
        & Set $k_\mathrm{max}^\text{discard}$ as the GT cardinality.
        \tabularnewline \hline
        Start- and End-points 
        & Require every motif to start and end around the start- and end-points of the GT motifs.
        & Use constraints~7--8 from the catalogue with the mask being 1 only within $\Delta l$ from the GT motifs' start- and end-points, where $\Delta l$ is $1/4$ of the average motif length.
        \tabularnewline \hline
        Non-consecutive Motifs 
        & Require a minimum time duration between motifs (within or across motif sets).
        & Define $\hmotsdiff(\beta, \beta') = \hmotssame(\beta, \beta') \;:\;\neg (b \leq b' \leq e + l_\text{buffer}) \;\land\; \neg(b' \leq b \leq e' + l_\text{buffer})$ where $l_\text{buffer}$ is half of the average time duration between consecutive motifs in the time series.
        \tabularnewline \hline
        Positive Region 
        & There has to be at least one motif in a specified `positive' region.
        & Select a random GT motif $[b:e]$, extend it into a positive region as $\theta = {[b-\frac{e-b+1}{2}:e+\frac{e-b+1}{2}]}$, and set $\hmset(\mset) \;:\; ( \exists \beta \in \mset : \beta \subseteq \theta )$.
        \tabularnewline \hline
        Length 
        & Require all motifs to be within the length range.
        & Use constraint~5 from the catalogue with $(l_{\min},l_{\max})$ being the length range of the GT motifs.
        \tabularnewline \hline
    \end{tabular}
    \caption{Hard constraints for every GT motif set in the benchmark data.} 
    \label{tab:hard_constraints_for_benchmark}
\end{table}

\begin{table}[tb]
    \centering
    \scriptsize
    \begin{tabular}{L{1.75cm}L{3.25cm}L{6.75cm}}
        \hline
        \textbf{Constraint Type} 
        & \textbf{Description} 
        & \textbf{Implementation} \newline (See Table~\ref{tab:example_constraints} for the catalogue.)
        \tabularnewline \hline 
        Cardinality 
        & Cardinalities closer to the specified cardinality are more desirable.
        & Use constraints~1--2 from the catalogue with $k_{\min}$ and $k_{\max}$ being the GT cardinality.
        \tabularnewline \hline 
        Soft Mask 
        & Motifs that occur in the specified regions are more desirable.
        & Specify a mask $\mathbf{m}$ that is 1 during GT motifs and 0 elsewhere. Then, `soften' the mask by making every transition linear with a duration of half of the average motif length (centered at the time of the original transition). Set the desirability of a motif set as the average of the mask values covered by the motifs: 
        $\dmset(\mset) = \avg (\textbf{m}_\mathcal{I})$ with $\mathcal{I} = \textstyle\bigcup_{\beta \in \mset} \beta$.
        \tabularnewline \hline 
        Positive Region
        & It is desirable to have at least one motif that is fully in the specified positive region.
        & Determine the positive region $\theta$ as in the corresponding hard constraint. For every motif $\beta \in \mset$, calculate the ratio of its time indices that are in $\theta$ as $r(\beta) = \frac{\beta \cap \theta}{\beta}$. Set the desirability to the maximum of these ratios: $\dmset(\mset) = \max_{\beta \in \mset} r(\beta)$. %
        \tabularnewline \hline 
        Length 
        & It is more desirable to have motifs within the length range.
        & Calculate the desirability as the ratio of the motifs that are in the GT length range.
        \tabularnewline \hline
    \end{tabular}
    \caption{Soft constraints for every GT motif set in the benchmark data.} 
    \label{tab:soft_constraints_for_benchmark}
\end{table}

A possible reason for the adverse effect of the Length-related constraints i
s
\locomotifdok's time warping capability, which causes candidate motif sets to consist of possibly variable-length motifs, some of which may not be admissible or desired according to constraints. 
This problem could be solved if \locomotifdok could compute time warping by considering the constraints; however, this is not straightforward to achieve in a computationally tractable manner, especially while supporting generic constraints.

Overall, domain knowledge is more helpful for variable-length datasets than fixed-length ones. 
This is because motif (set)s with more diverse properties enable the domain knowledge to be more specialized for individual motif sets, which can be successfully leveraged by \locomotifdok.

To conclude, the majority of the constraints we consider lead to higher-quality motif sets; hence, our response to RQ2 is `most of the time'.

\subsection{Validation Based on Selected Time Series}\label{sec:validation}

Leveraging domain knowledge represented by a mask $\mathbf{m}$ has been shown to be essential for the existing TSMD technique MPV to discover interesting motifs in various applications~\citep{guiding}.
This leads us to:
\begin{enumerate}[font=\bfseries, wide=0pt]
    \item[RQ3:] Can \locomotifdok leverage the available domain knowledge %
    in multiple applications to discover (only) interesting motifs?
\end{enumerate}
To answer this question, we consider the time series selected by~\cite{guiding} from different application domains and
evaluate the results qualitatively since there are no GT motif sets available. 
We present the results for the ECG time series
in this section and for the remaining applications in Appendix~\ref{appendix:additional_example_time_series}.

\begin{figure}[tb]
    \centering
    %
    % \includesvg[width=\textwidth]{figures/ecg_5_paper_fig_updated.svg}
    \includegraphics[width=\textwidth]{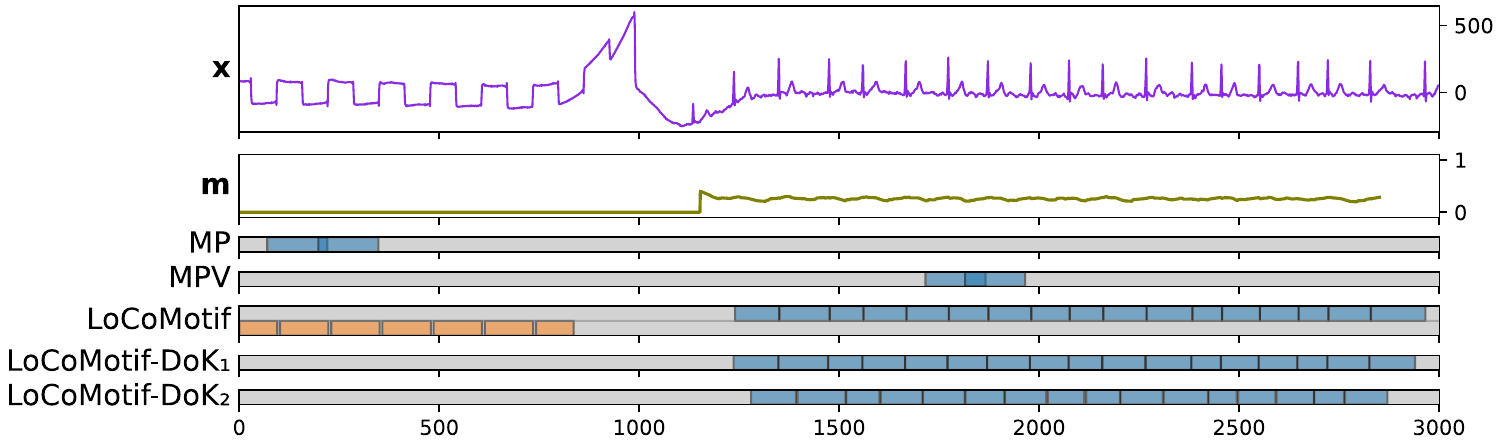}
    \\[.5cm]
    % \includesvg[width=\textwidth]{figures/ecg_5_paper_fig2_updated.svg}
    \includegraphics[width=\textwidth]{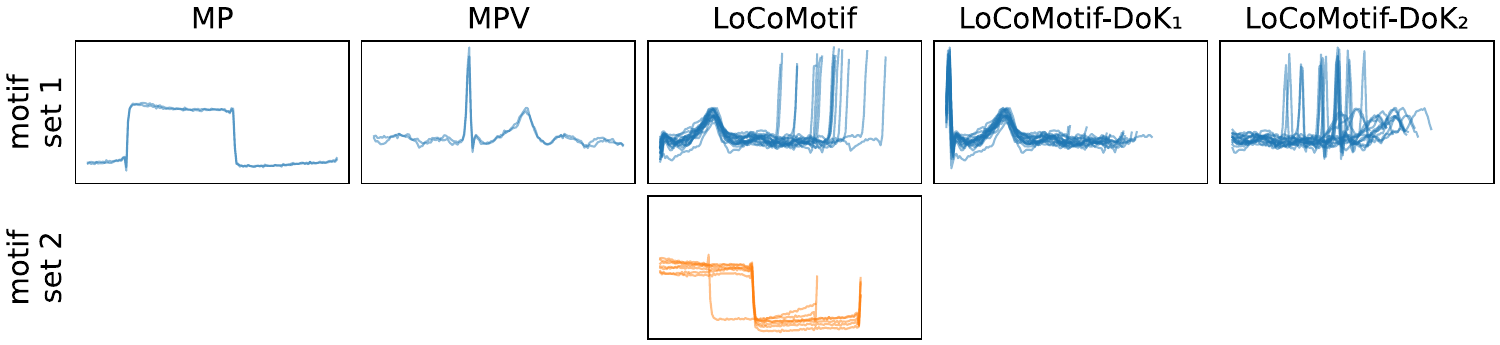}
    \caption{%
    In the ECG time series $\mathbf{x}$, domain knowledge is represented by a mask $\mathbf{m}$ that quantifies the desirability of motifs that start at every time index. 
    TSMD techniques MP and LoCoMotif, which cannot leverage $\mathbf{m}$, discover uninteresting calibration waves as motifs. 
    In contrast, MPV and $\text{\locomotifdok}_1$ discover only heartbeats as motifs, as desired, by leveraging $\mathbf{m}$.
    $\text{\locomotifdok}_2$ uses additional domain knowledge to discover motifs where the peaks of heartbeats are roughly centered.
    }
    \label{fig:ecg}
\end{figure}

The ECG time series $\mathbf{x}$ plotted in Figure~\ref{fig:ecg} consists of a calibration signal and heartbeats~\cite{petrutiu2007}, and only the latter are interesting to discover as motifs. 
The figure also shows the motifs discovered by different TSMD techniques as time segments (filled rectangles) and overlaid subsequences. 

Among the existing TSMD techniques that don't leverage domain knowledge, 
Matrix Profile (MP) discovers a pair of calibration waves, and LoCoMotif discovers two motif sets, one consisting of heartbeats and the other calibration waves.\footnote{%
For MP and MPV, we use the motif length $l=150$, as in~\cite{guiding}. 
For LoCoMotif(-DoK), we set $(l_{\min}, l_{\max}) = (0.5l, 1.5l)$, $\rho=0.6$, and $\nu=0$. 
} 
While the outcome of MP is not useful at all, the result of LoCoMotif is satisfactory, as one of the two discovered motif sets is interesting. 
LoCoMotif obtains a satisfactory result due to its ability to discover multiple motif sets and its consideration of the total duration of motifs in a motif set (in addition to the similarity between them) which favors more frequently occurring heartbeats over calibration waves.

The essential domain knowledge in this ECG application can be expressed as \textit{``motifs should not be similar to the calibration signal.''} 
We represent this knowledge in the same way as~\cite{guiding}, by a mask $\mathbf{m}$ that quantifies the desirability of motifs that start at every time index. 
For the example time series, Figure~\ref{fig:ecg} shows $\mathbf{m}$, which disallows motifs that start before the time index 1153, after which the desirability fluctuates. 
Leveraging $\mathbf{m}$ improves the discovered motifs: 
MPV identifies two heartbeats and \locomotifdok (denoted as $\text{\locomotifdok}_1$) discovers all the heartbeats without discovering the calibration signal as another motif set.

The subsequences in Figure~\ref{fig:ecg} show that motifs discovered by $\text{\locomotifdok}_1$ do represent periods of the quasiperiodic heartbeat signal; however, these motifs start with the peaks while it might be more preferable to have the peaks centered in time. 
To this end, we define additional domain knowledge through two constraints: 
First, we define a hard constraint that requires representative motifs to have a minimum skewness of $\gamma_{\min} = 2.5$ as 
$\hmotrepr(\alpha) \;:\;  \text{skewness}{( \mathbf{x}_\alpha )} \geq \gamma_{\min}$.
because a subsequence that contains a peak has a positively skewed distribution of measurements. 
Second, we define a soft constraint where the desirability of a motif set is 1 when its representative's peak (i.e., the maximum value) is centered in time, and linearly decreases to 0 when the peak approaches the beginning or the end: 
\begin{equation}
    \dmset(\{ [b_1 : e_1], \ldots, [b_k : e_k] \}) = 
    \begin{cases}
        \frac{2i_\text{peak}}{e_1 - b_1}  & \text{if $i_\text{peak} \leq \frac{e_1 - b_1}{2}$} \\ 
        2 - \frac{2i_\text{peak}}{e_1 - b_1}  & \text{otherwise}
    \end{cases}
\end{equation}
where $i_\text{peak} = \argmax{\mathbf{x}_{[b_1 : e_1]}}$ is the time index of the peak within the representative motif $\mathbf{x}_{[b_1 : e_1]}$. 
This constraint favors motifs where the peak occurs at the center (if any). 
$\text{\locomotifdok}_2$ represents the incorporation of these additional constraints in Figure~\ref{fig:ecg}. 
Unlike $\text{\locomotifdok}_1$, $\text{\locomotifdok}_2$ discovers motifs that contain heartbeats with the peaks being roughly centered, as desired.

Therefore, the answer to RQ3 is yes: \locomotifdok successfully leverages the type of domain knowledge considered by MPV. 
While domain knowledge is essential for MP(V) to discover the desired pattern (heartbeats), it is useful for LoCoMotif(-DoK) to discover only the desired pattern. 
Moreover, \locomotifdok can leverage additional domain knowledge represented by constraints to discover the desired period of the quasiperiodic heartbeat signal.
Domain knowledge is useful in the remaining applications as well, which are presented in Appendix~\ref{appendix:additional_example_time_series}.

\section{Conclusion}\label{sec:conclusion}
We have presented a framework that incorporates domain knowledge into TSMD in a generic form and an efficient implementation of the framework with the \locomotifdok algorithm. 
We have quantitatively and qualitatively evaluated \locomotifdok in a real physiotherapy use case, on a TSMD benchmark, and on other time series from different application domains.
We have shown that \locomotifdok successfully leverages various types of domain knowledge and discovers higher-quality motifs than existing TSMD techniques in most cases. 

As future work, one might optimize the efficiency of \locomotifdok for certain constraints.

\noindent \textbf{Acknowledgements} This research received funding from the Flemish Government under the ``Onderzoeksprogramma Artificiële Intelligentie (AI) Vlaanderen'' programme and the VLAIO ICON-AI Conscious project (HBC.2020.2795).

% \bibliography{references}

\clearpage

\appendix
\section{Additional Example Time Series for Validation}\label{appendix:additional_example_time_series}

Here we answer RQ2 in Section~\ref{sec:validation} for the four remaining example time series from~\cite{guiding}. 
We use the same experimental setup as the ECG example in Section~\ref{sec:validation} %
and the same domain knowledge as the original paper~\citep{guiding}.
The motif length hyperparameter $l$ of MP(V) is selected in the same way as~\cite{guiding}. 
For LoCoMotif(-DoK), we select $(l_{\min},l_{\max})$ according to $l$, disable time warping, and select a reasonable strictness ($\rho$). 
Hyperparameter values are provided in the captions of the figures that present the results.

\begin{figure}[h]
    \centering
    \vspace{-0.5cm}
    % \includesvg[width=\textwidth]{figures/eog_3_paper_fig_updated.svg}
    \includegraphics[width=\textwidth]{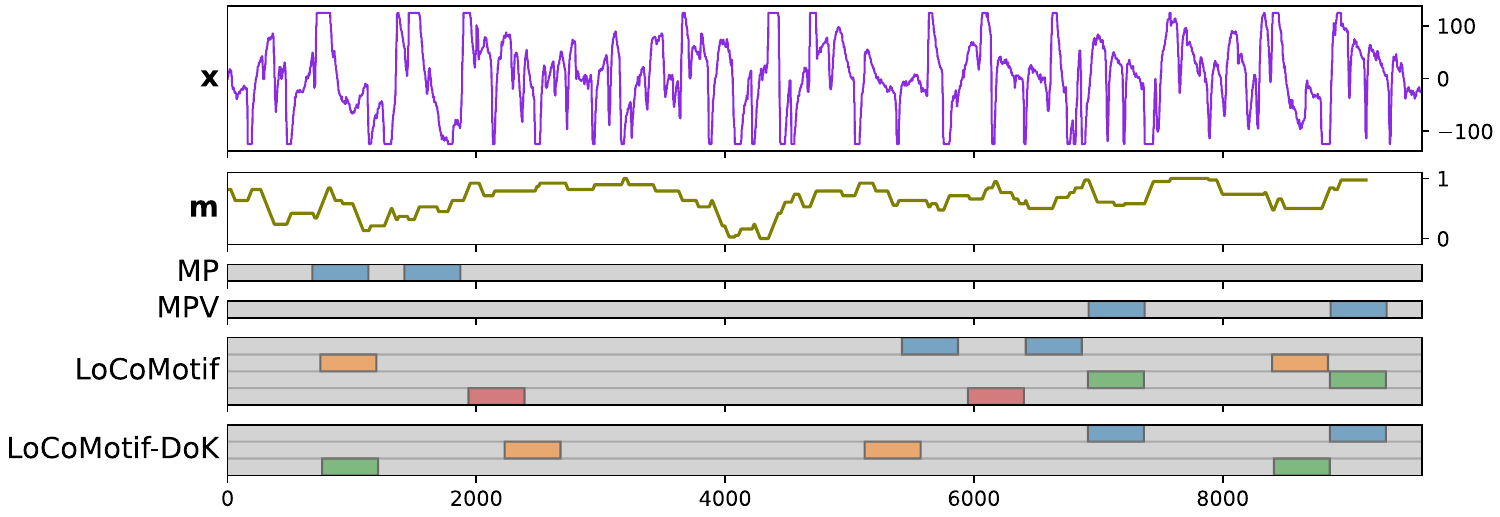}
    \\[.5cm]
    % \includesvg[width=\textwidth]{figures/eog_3_paper_fig2_updated.svg}
    \includegraphics[width=\textwidth]{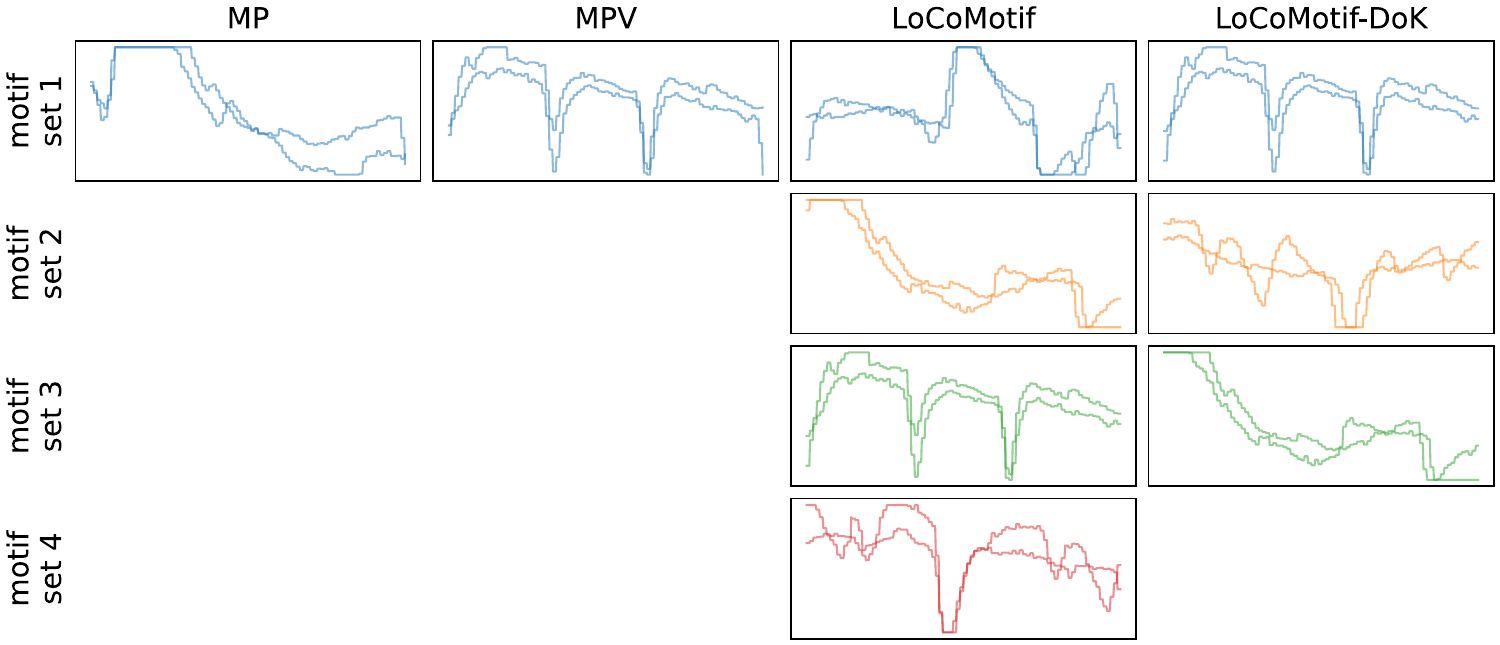}
    \caption{%
    The EOG example. 
    Hyperparameters: $l=l_{\min}=l_{\max}=450$, $\rho=0.4$.
    }
    \label{fig:eog}
\end{figure}

The \textbf{electrooculoram (EOG)} time series that captures eye movements suffers from sensor saturation, which causes flat plateaus that are similar to each other (Figure~\ref{fig:eog}) and don't have any medical significance. 
The domain knowledge is \textit{``motifs should not include saturated measurements''} and is represented as a mask $\mathbf{m}$ based on the ratio of unsaturated measurements over a sliding window. 
By leveraging $\mathbf{m}$, the MPV technique discovers a medically interesting motif pair instead of the regions with sensor saturation (discovered by MP). 
The interesting motif pair is discovered as the third motif set by LoCoMotif and as the first motif set by \locomotifdok thanks to the domain knowledge. 
Hence, \locomotifdok performs better than LoCoMotif, considering that the motif sets discovered by LoCoMotif(-DoK) are sorted from the best to the worst (unless constraints are determined differently in \locomotifdok for every (pair of) motif set(s) to be discovered, which is not the case here).

\begin{figure}[h]
    \centering
    \vspace{-0.5cm}
    % \includesvg[width=\textwidth]{figures/fnirs_2_paper_fig_updated.svg}
    \includegraphics[width=\textwidth]{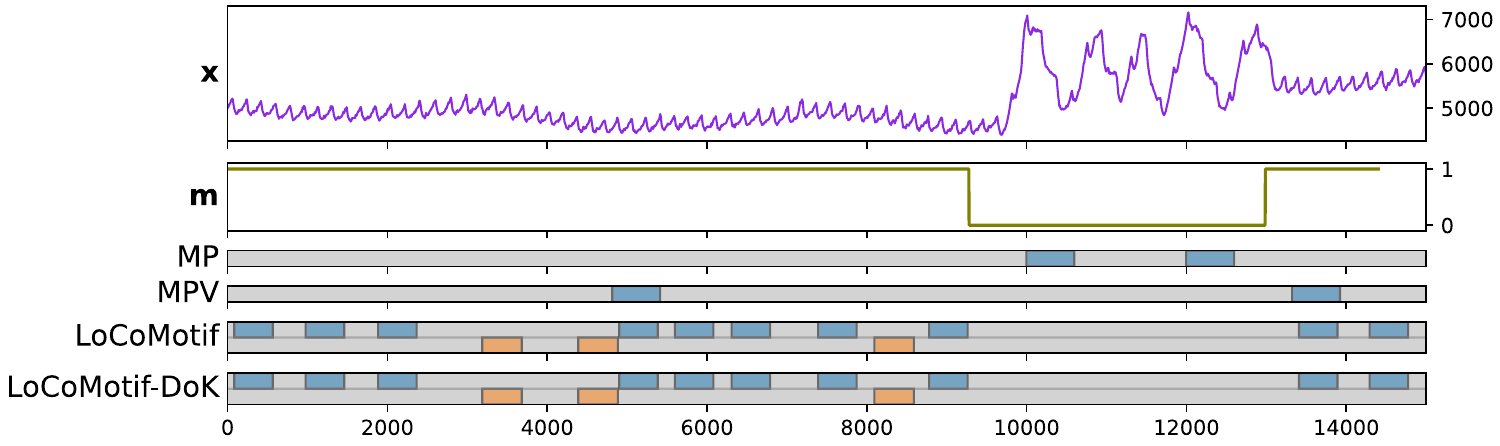}
    \\[.5cm]
    % \includesvg[width=\textwidth]{figures/fnirs_2_paper_fig2_updated.svg}
    \includegraphics[width=\textwidth]{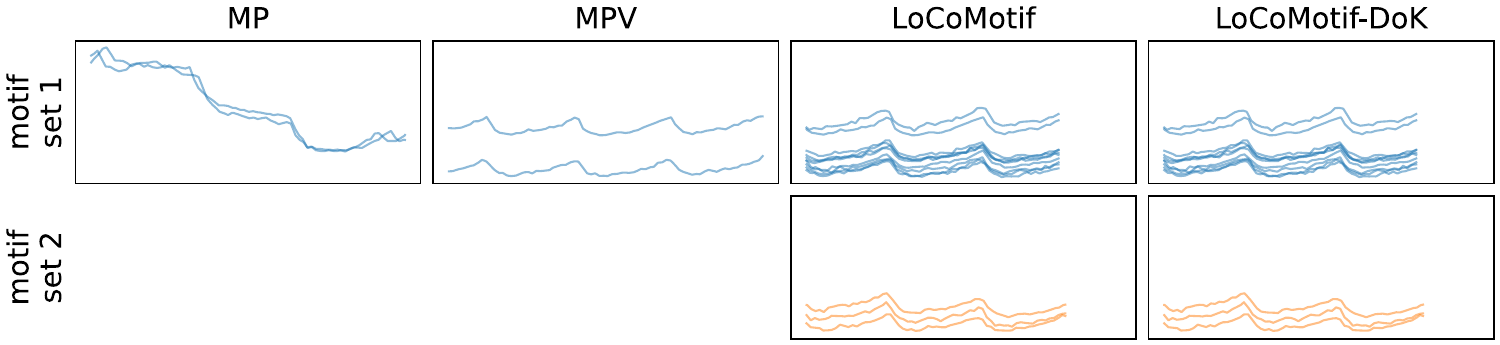}
    \caption{%
    The fNIRS example.
    Hyperparameters: $l=600$, $(l_{\min},l_{\max})=(0.75l,1.25l)$, $\rho=0.3$.
    }
    \label{fig:fnirs}
\end{figure}

The \textbf{Functional Near Infrared Spectroscopy (fNIRS)} data that represents brain activities contains sensor artifacts due to motion of the body (high-amplitude fluctuations in Figure~\ref{fig:fnirs}). 
The domain knowledge is \textit{``motifs must not be found when there is substantial body motion''}.
To this end, time segments with substantial motion are detected based on another sensor (a body-worn accelerometer) and the corresponding values of $\mathbf{m}$ are set to 0 while the remaining are 1. 
The use of $\mathbf{m}$ enables MPV to discover a pair of medically significant motifs that represent neuronal activities instead of sensor artifacts (discovered by MP). 
On the other hand, both LoCoMotif and \locomotifdok obtain desired results: they only discover medically significant motifs, regardless of whether domain knowledge is used or not.

\begin{figure}[h]
    \centering
    \vspace{-0.5cm}
    % \includesvg[width=\textwidth]{figures/finger_flexion_3_paper_fig_updated.svg}
    \includegraphics[width=\textwidth]{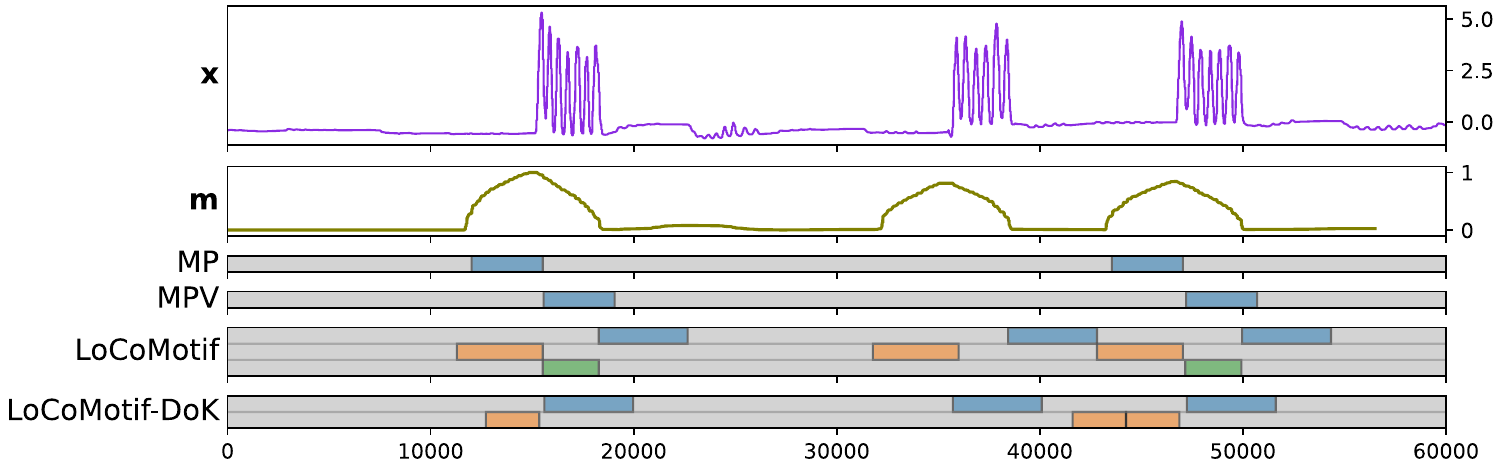}
    \\[.5cm]
    % \includesvg[width=\textwidth]{figures/finger_flexion_3_paper_fig2_updated.svg}
    \includegraphics[width=\textwidth]{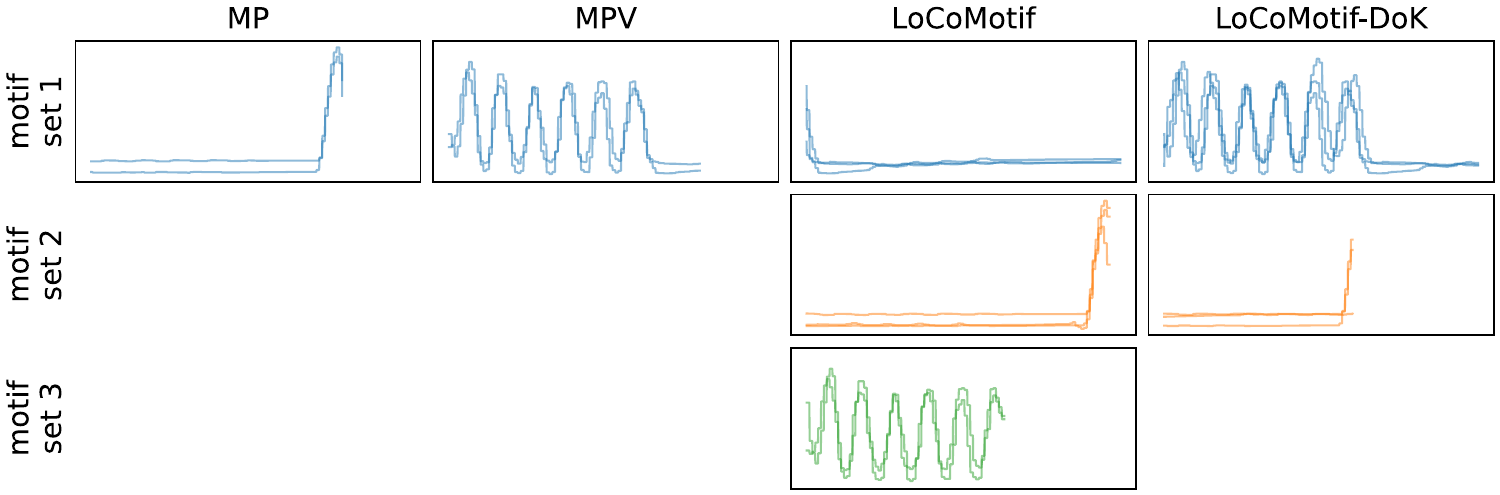}
    \caption{%
    The ECoG example.
    Hyperparameters: $l=3500$, $(l_{\min},l_{\max})=(0.75l,1.25l)$, $\rho=0.1$.
    }
    \label{fig:ecog}
\end{figure}

The \textbf{electrocorticography (ECoG)} time series measures the flexion of a finger. The example time series in Figure~\ref{fig:ecog} contains three flexions, which causes three time periods with high-frequency oscillations that are interesting to discover. 
The domain knowledge in this application is \emph{``not discovering simple motifs''}.
This is represented by $\mathbf{m}$ that is calculated based on the ``complexity'' of the time series over a sliding window, where the complexity of a subsequence is measured by the Euclidean norm of the first-order difference. 
MP-based algorithms need the domain knowledge to discover the interesting finger flexions as motifs (although they can discover only two of them by design). 
Without domain knowledge, LoCoMotif obtains a satisfactory result: it discovers three motif sets where the last one consists of two finger flexions. 
\locomotifdok outperforms LoCoMotif thanks to the domain knowledge by successfully discovering all the three finger flexions as the first motif set and some uninteresting patterns as the second motif set. 
We note that $\mathbf{m}$ approaches 0 but does not reach it most of the time; hence, motifs are allowed in these regions, albeit highly discouraged. 
    
\begin{figure}[h]
    \centering
    \vspace{-0.5cm}
    % \includesvg[width=\textwidth]{figures/seismology_2_paper_fig_updated.svg}
    \includegraphics[width=\textwidth]{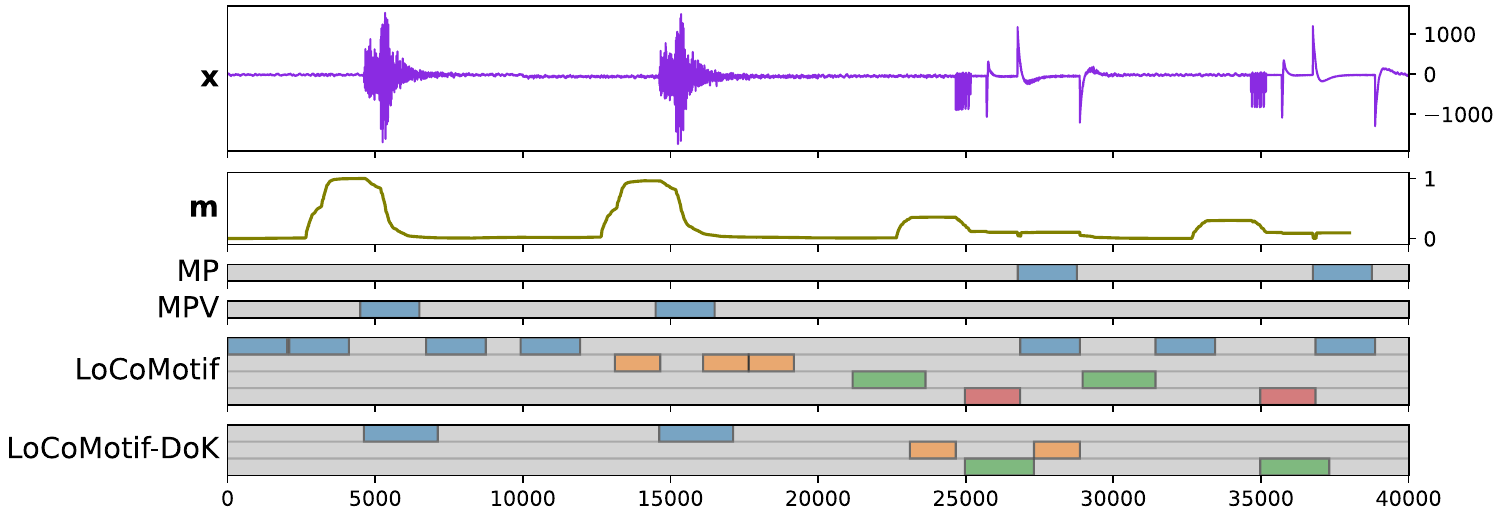}
    \\[.5cm]
    % \includesvg[width=\textwidth]{figures/seismology_2_paper_fig2_updated.svg}
    \includegraphics[width=\textwidth]{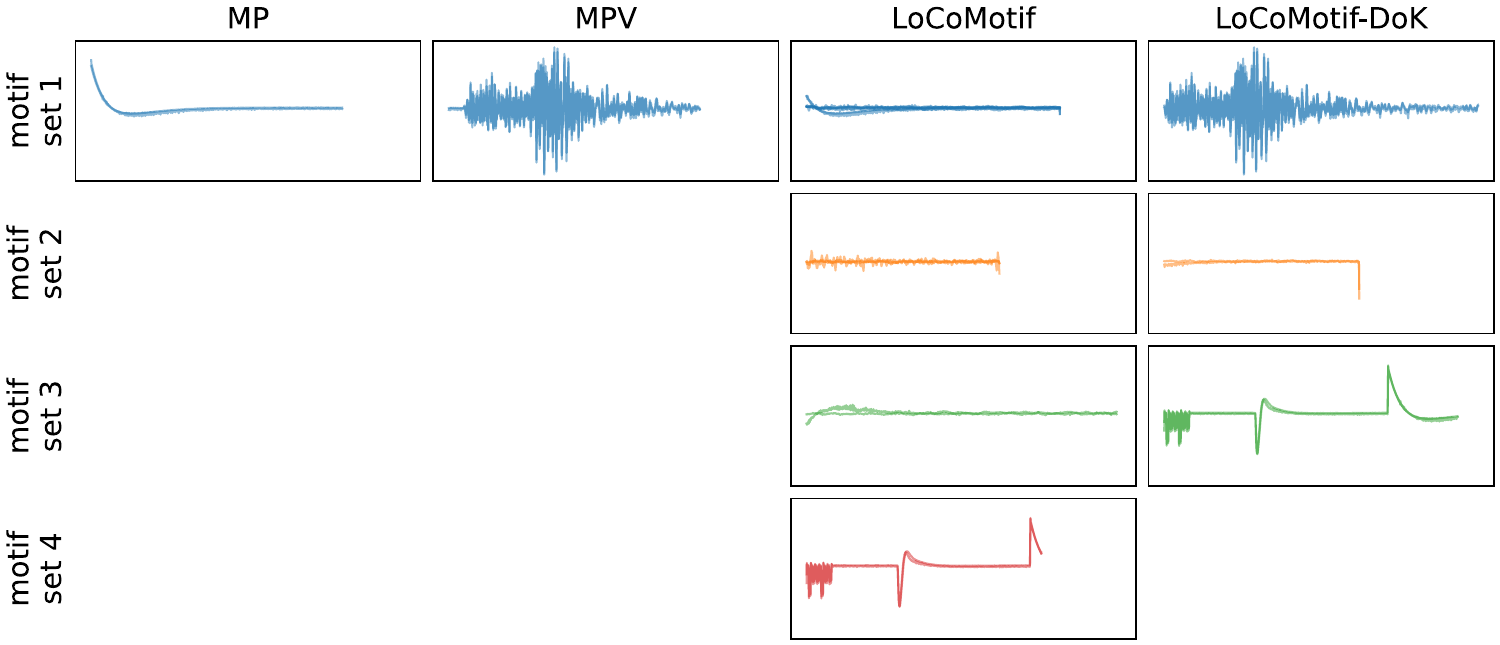}
    \caption{%
    The seismology example.
    Hyperparameters: $l=2000$, $(l_{\min},l_{\max})=(0.75l,1.25l)$, $\rho=0.1$.
    }
    \label{fig:seismology}
\end{figure}

The \textbf{seismology} data shown in Figure~\ref{fig:seismology} contains two earthquakes (around time samples 5,000 and 15,000) that are interesting to discover. 
However, neither MP nor LoCoMotif can discover them because they cannot leverage the domain knowledge. 
As in the ECoG application, the domain knowledge is represented by $\mathbf{m}$ calculated based on the complexity. 
Thanks to the use of $\mathbf{m}$, both MPV and \locomotifdok discover the two earthquakes, \locomotifdok additionally discovering other patterns as another two motif sets. 

From these four examples, we draw a similar conclusion to the ECG example in Section~\ref{sec:validation}. 
\locomotifdok successfully leverages the domain knowledge defined for MPV in multiple applications. 
In general, domain knowledge is necessary for MP(V) to discover interesting motifs, whereas it is either necessary or useful for LoCoMotif(-DoK), depending on the application.

\end{document}